\newif\iftodo
\todotrue

\documentclass[journal]{IEEEtran}
\usepackage{times}
\usepackage{cite}
\usepackage{multicol}
\usepackage[hidelinks]{hyperref}
\usepackage{graphicx} % for pdf, bitmapped graphics files
\usepackage{epsfig} % for postscript graphics files
\usepackage{times} % assumes new font selection scheme installed
\usepackage{amsmath} % assumes amsmath package installed
\usepackage{amssymb}  % assumes amsmath package installed
\let\labelindent\relax
\usepackage{enumitem}
\usepackage[acronym]{glossaries}
\usepackage[normalem]{ulem}   
\usepackage{color}
\usepackage[ruled,linesnumbered]{algorithm2e}
\usepackage{mathtools}
\usepackage{array}
\usepackage{dblfloatfix}

\usepackage{amsthm}
\newtheorem{example}{Example}

\makeatletter
\newcommand{\removelatexerror}{\let\@latex@error\@gobble}
\makeatother

\definecolor{purple}{rgb}{0.5,0,0.5}
\iftodo
\newcommand{\adam}[1]{{\textcolor{purple}{[Adam] \emph{\bf#1 }}}}
\newcommand{\hkc}[1]{{\textcolor{blue}{[HKG]\emph{\bf#1 }}}} 
\newcommand{\hkg}[1]{{\textcolor{ magenta}{#1}}} 
\newcommand{\himani}[1]{{\textcolor{cyan}{[Himani] \emph{\bf#1 }}}}
\else
\newcommand{\adam}[1]{}
\newcommand{\hkc}[1]{}
\newcommand{\hkg}[1]{}
\newcommand{\himani}[1]{}
\fi
\newenvironment{flushitemize}{
\begin{list}{$\bullet$}
   {\setlength{\leftmargin}{15pt}}
    \setlength{\labelwidth}{20pt}
    \setlength{\itemindent}{0pt}
    \setlength{\labelsep}{0.5em}
 \setlength{\itemsep}{1pt}
 \setlength{\parskip}{0pt}
 \setlength{\parsep}{0pt}}
 {\end{list}}

\newacronym{dfa}{DFA}{Deterministic Finite Automaton}
\newacronym{sccs}{SCCs}{Strongly Connected Components}
\newacronym{mdp}{MDP}{Markov Decision Process}
\newacronym{fts}{FTS}{Finite Transition System}
\newacronym{ltl}{LTL}{Linear Temporal Logic}
\newacronym{pctl}{PCTL}{Probabilistic Computation Tree Logic}
\newacronym{wdfa}{WDFA}{Weighted Deterministic Finite Automaton}
\newacronym{scc}{SCC}{Strongly Connected Component}
\newacronym{scltl}{scLTL}{Co-Safe Linear Temporal Logic}
\newacronym{gr1}{GR(1)}{Generalized Reactivity (1)}
\newacronym{dmp}{DMP}{Dynamic Motion Primitive}
\newacronym{dmps}{DMPs}{Dynamic Motion Primitives}
\newacronym{tamp}{TAMP}{Task and Motion Planning}

\newcommand{\vars}{\mathcal{V}}
\newcommand{\dom}{\Sigma}
\newcommand{\game}{\mathcal{G}}
\newcommand{\grounding}{\textrm{G}}
\newcommand{\winning}{Z}
\newcommand{\initstates}{\theta^\textrm{init}}
\newcommand{\obj}{\Phi}

\newcommand{\inp}{\mathcal{I}}
\newcommand{\inpsyms}{\mathcal{I}_\grounding}
\newcommand{\inpuser}{\mathcal{I}_\textrm{u}}
\newcommand{\out}{\mathcal{O}}
\newcommand{\inpstate}{\statevar_{\inp}}
\newcommand{\outstate}{\statevar_{\out}}
\newcommand{\statevar}{\sigma}
\newcommand{\var}{\pi}
\newcommand{\inpstates}{\dom_\inp}
\newcommand{\outstates}{\dom_\out}
\newcommand{\inpstatesprime}{\dom_{\inp'}}

\newcommand{\inpstatesdoubleprime}{\dom_{\inp^{\dagger}}}
\newcommand{\allstates}{\dom_\vars}
\newcommand{\allstatesprime}{\dom_{\vars'}}

\newcommand{\inpstateone}{\statevar_{\inp, \textrm{1}}}
\newcommand{\inpstateonea}{\statevar_{\inp, \textrm{1,A}}}
\newcommand{\inpstatetwo}{\statevar_{\inp, \textrm{2}}}
\newcommand{\inpstatetwoa}{\statevar_{\inp, \textrm{2,A}}}
\newcommand{\outstateone}{\statevar_{\out, \textrm{1}}}
\newcommand{\outstatetwo}{\statevar_{\out, \textrm{2}}}
\newcommand{\inpstatethree}{\statevar_{\inp, \textrm{3}}}
\newcommand{\inpstatethreea}{\statevar_{\inp, \textrm{3,A}}}
\newcommand{\outstatethree}{\statevar_{\out, \textrm{3}}}

\newcommand{\setor}{~|~}

\newcommand{\true}{{\tt{True}}}
\newcommand{\false}{{\tt{False}}}

\newcommand{\skills}{\mathcal{A}}
\newcommand{\skill}{\texttt{skill}}

\newcommand{\trans}{\tau}
\newcommand{\tenv}{\trans_\textrm{e}}
\newcommand{\tenvhard}{\trans_\textrm{e}^\textrm{hard}}

\newcommand{\tsys}{\trans_\textrm{s}}

\newcommand{\tsyshard}{\trans_\textrm{s}^\textrm{hard}}
\newcommand{\tsysnothard}{\trans_\textrm{s}^\textrm{mutable}}
\newcommand{\tenvnew}{\trans_\textrm{e}^\textrm{new}}

\newcommand{\tfullskills}{\trans_\textrm{skills}}
\newcommand{\tnotwinningfullskills}{\trans_\textrm{not-winning}}
\newcommand{\tenvadd}{\trans_\textrm{env-add}}
\newcommand{\tenvold}{\trans_\textrm{env-old}}

\newcommand{\tsyscanwin}{\trans_\textrm{sys-can-win}}
\newcommand{\tsysalwayswins}{\trans_\textrm{sys-always-wins}}

\newcommand{\treachable}{\trans_\textrm{reachable}}
\newcommand{\tcanwinandencoded}{\trans_\textrm{always-win-and-reachable}}
\newcommand{\tpreskills}{\trans_\textrm{pre-skills}}
\newcommand{\tfinalpost}{\trans_\textrm{final-post}}
\newcommand{\tnotallowedrepair}{\trans_\textrm{disallowed-transitions}}
\newcommand{\tchangeconstraints}{\trans_\textrm{poss-changes}}
\newcommand{\tchangecons}{\trans_\textrm{poss-changes}}
\newcommand{\tallowablechanges}{\trans_\textrm{allowable-changes}}
\newcommand{\tselectedchange}{\trans_\textrm{selected-change}}
\newcommand{\tnewpreprecondition}{\trans_\textrm{new-pre-precondition}}
\newcommand{\tenvnewfullskill}{\trans_\textrm{env-new-skill}}
\newcommand{\tnewfullskill}{\trans_\textrm{new-skill}}
\newcommand{\toldfullskill}{\trans_\textrm{old-skill}}
\newcommand{\tenvprenew}{\trans_\textrm{env-pre-new}}
\newcommand{\tenvpreold}{\trans_\textrm{old-pre-new}}
\newcommand{\toldprepreconditions}{\trans_\textrm{old-pre-preconditions}}

\newcommand{\tsysnewmutable}{\tau_s^\textrm{new-mutable}}

\newcommand{\envspec}{\varphi_\textrm{e}}
\newcommand{\sysspec}{\varphi_\textrm{s}}
\newcommand{\envinit}{\varphi_\textrm{e}^\textrm{i}}
\newcommand{\sysinit}{\varphi_\textrm{s}^\textrm{i}}
\newcommand{\envsafety}{\varphi_\textrm{e}^\textrm{t}}
\newcommand{\envsafetynothard}{\varphi_\textrm{e}^\textrm{t,mutable}}
\newcommand{\envsafetyhard}{\varphi_\textrm{e}^\textrm{t,hard}}
\newcommand{\envsafetyharduser}{\varphi_\textrm{e}^\textrm{t,user}}
\newcommand{\syssafety}{\varphi_\textrm{s}^\textrm{t}}
\newcommand{\syssafetynothard}{\varphi_\textrm{s}^\textrm{t,mutable}}
\newcommand{\syssafetyhard}{\varphi_\textrm{s}^\textrm{t,hard}}
\newcommand{\syssafetyuser}{\varphi_\textrm{s}^\textrm{t,user}}
\newcommand{\syssafetycontinue}{\varphi_\textrm{s}^\textrm{t,continue}}
\newcommand{\syssafetychoose}{\varphi_\textrm{s}^\textrm{t,choose}}
\newcommand{\envlive}{\varphi_\textrm{e}^\textrm{g}}
\newcommand{\syslive}{\varphi_\textrm{s}^\textrm{g}}
\newcommand{\changecons}{\varphi_\textrm{poss-changes}}
\newcommand{\notallowedrepair}{\varphi_\textrm{disallowed-transitions}}

\glsdisablehyper

\begin{document}

\title{Physically-Feasible Repair of Reactive, Linear Temporal Logic-based, High-Level Tasks}

\author{Adam Pacheck and Hadas Kress-Gazit~\IEEEmembership{Fellow,~IEEE}
        % <-this % stops a space
\thanks{The authors are with Cornell University in Ithaca, NY 14850, USA. Email: \{akp84, hadaskg\}@cornell.edu. This work has been submitted to the IEEE for possible publication. Copyright may be transferred without notice, after which this version may no longer be accessible.}% <-this % stops a space
%\thanks{Manuscript received April 26, 2022}
}

\markboth{Transactions on Robotics,~Vol.~XX, No.~XX, Month~202X}%
{Physically-Feasible Repair of Reactive, Linear Temporal Logic-based, High-Level Tasks}

\IEEEpubid{0000--0000/00\$00.00~\copyright~2021 IEEE}

\maketitle

\begin{abstract}
A typical approach to creating complex robot behaviors is to compose atomic controllers, or skills, such that the resulting behavior satisfies a high-level task; however, when a task cannot be accomplished with a given set of skills, it is difficult to know how to modify the skills to make the task possible. We present a method for combining symbolic repair with physical feasibility-checking and implementation to automatically modify existing skills such that the robot can execute a previously infeasible task.

We encode robot skills in \gls{ltl} formulas that capture both safety constraints and goals for reactive tasks. Furthermore, our encoding captures the full skill execution, as opposed to prior work where only the state of the world before and after the skill is executed are considered. Our repair algorithm suggests symbolic modifications, then attempts to physically implement the suggestions by modifying the original skills subject to \gls{ltl} constraints derived from the symbolic repair. If skills are not physically possible, we automatically provide additional constraints for the symbolic repair. We demonstrate our approach with a Baxter and a Clearpath Jackal.

\end{abstract}


\section{Introduction}
\glsresetall
A central challenge in robot planning and control is how to automatically and correctly generate robot behavior that satisfies complex, high-level tasks, especially those where the robot must react to events in the environment (i.e. \textit{reactive} tasks).
One common approach is to compose skills--atomic behaviors--that are simpler to define and verify than the full complex behavior.
For example, for a table setting robot, it is simpler to provide the robot with individual skills for placing  silverware, plates, and cups, and define tasks for performing different place settings, than to create one ``set table'' skill.

There are multiple approaches to skill composition, from chaining skills/controllers together \cite{konidaris2018skills, tedrake2010lqr, majumdar2017funnel, burridge1999} to composing them based on symbolic controllers synthesized from temporal logic~\cite{kress2018synthesis}.
Typically, such approaches abstract skills and states as symbols, then synthesize a symbolic high-level strategy that is used to govern the execution of the skills by the physical system, based on environment or user-provided input. 
While powerful, such approaches traditionally make two assumptions that may not hold; first, that skills are abstracted by defining preconditions and postconditions (e.g. \cite{konidaris2018skills, lavalle2006planning, fikes1971strips})--states in which the skill can be executed and states that are the result of applying the skill, respectively--while in fact, during skill execution the system might be temporarily in another state. 
Second, that all the skills needed to achieve the task are given. 
In this work we relax those two assumptions.

For the first assumption, without considering what happens during the execution of the skill, the tasks we can specify and verify are limited--we cannot verify safety constraints or goals that might be reached during the execution of the skill.
For example, we might not want the table setting robot to move drinks over any electronics on the table, but given only the start location of the drink and the end location (state before and after the skill execution), we might think the constraint is satisfied, while in reality it is not. In this work we explicitly consider such \textit{intermediate states}, thereby ensuring correct \textit{physical} execution of the task. 

To relax the second assumption, ideally the system would be able to recognize and repair tasks that cannot be achieved with the set of skills the robot currently has. 
While there exist several methods to repair temporal logic specifications (e.g. \cite{pacheck2020finding, fainekos2011revising, alur2013counter}), provide reasons for infeasibility (e.g. \cite{raman2013explaining}), or learn skills subject to constraints (e.g. \cite{innes2020elaborating, rana2018towards, toro2018teaching}), there exists little work on automatically finding \textit{physically implementable} skills to repair a  high-level task.
In this work, we integrate symbolic repair of high-level tasks with physical feasibility-checking and implementation of skills that were symbolically suggested.

In this work, we are given a reactive task encoded in the \gls{gr1} fragment of \gls{ltl} \cite{bloem2012synthesis}, a set of skills, and an abstraction of the world.
We encode the skills in a GR(1) formula such that intermediate states are explicitly encoded. 
Given the task, we automatically find a strategy that enables the robot to complete it, if possible,  or repair the task by finding and physically implementing skills that enable accomplishing the task.

During the repair process, not all skill suggestions will be physically possible.
For example, in the case of a table setting robot, a specific configuration of plates and cups may make a reaching skill impossible to physically execute without knocking over a cup.
We therefore divide the task repair process into two parts: symbolic repair---finding, symbolically, suggestions for skill modifications---and physical feasibility checking and implementation---finding a controller able to physically implement the modification.

Crucial and unique to our approach is the \textit{feedback} between the two parts; the symbolic repair provides suggestions to the physical repair, and the physical repair includes run-time monitors that, if a skill's repair is not physically feasible, provide additional constraints to the symbolic repair.
In addition, we \newpage\noindent  allow the user to provide constraints on how skills can be repaired, for example restricting the symbols that are allowed to change; these (optional) constraints can be used to guide the repair towards more physically desirable modifications.
Our repair process iterates between symbolic repair and physical implementation until a feasible physical skill, or set of skills, that repairs the overall task, is found.

\textbf{Contributions:} 
In this work, we provide a novel automated method for defining and repairing high-level, reactive behaviors given in \gls{gr1} such that they are physically feasible.
Our main contributions are: (1) Skill encoding that enables synthesizing control with physically-realistic safety constraints. 
(2) An automated process for physically implementing symbolic repair suggestions where the symbolic and physical repair are tightly integrated.
(3) Demonstrations of the approach with two different physical robots, abstractions, and tasks.

\section{Related Work}

\subsection{Linear Temporal Logic and Specification Repair}
\gls{ltl} has been used for specifying complex tasks in a variety of applications \cite{kress2018synthesis}.
However, if a task cannot be completed as specified, it is difficult to determine the cause, much less what must be done to make the task possible.
Recent work has investigated methods for debugging specifications that cannot be completed, leaving the user to determine the appropriate fix (e.g.~\cite{raman2013explaining, konighofer2009debugging}). 
In \cite{raman2013explaining}, the authors propose a method to highlight portions of an \gls{ltl} specification that make the task not possible.
Given a \gls{gr1} specification that cannot be satisfied, \cite{konighofer2009debugging} provides a countertrace that shows how the specification can be violated.

Researchers have investigated the automatic repair of specifications; these methods often suggest adding restrictions to the environment to prohibit behaviors that would make the specification unrealizable (e.g.~\cite{alur2013counter, fainekos2011revising, kim2015minimal, dokhanchi2015metric, chatterjee2008environment, li2011mining}).
Related to specification repair is the partial satisfaction of specifications, where a task is divided into hard constraints that must always hold and soft constraints that the robot does its best to satisfy (e.g.~\cite{lahijanian2012temporal, lahijanian2016iterative}).

This paper builds on recent work on specification repair that modifies the skills of the robot rather than restrict the environment \cite{pacheck2019automatic, pacheck2020finding}.
This work, in contrast to \cite{pacheck2019automatic} and \cite{pacheck2020finding}, explicitly considers the intermediate states of skills, thereby we are able to capture the full physical execution, not only the preconditions and postconditions as the previous works did.
There, multiple abstract new skills or modifications to skills were automatically suggested and then the user could choose which one(s) to try to physically implement on their own.
This work integrates the symbolic and physical repair by automatically implementing the suggested modification and provides additional constraints to the repair process if the symbolic suggestions are not physically possible.
The symbolic repair portion of this work additionally allows an entire new class of specifications to be repaired that \cite{pacheck2019automatic} and \cite{pacheck2020finding} can not, such as modifying the intermediate states of the table setting robot that moves drinks over electronics.

\subsection{Relation to LTL Constraints for Physical Skills}
Recent work has investigated incorporating \gls{ltl} specifications into physical skills.
Several works (e.g. \cite{li2017reinforcement, illanes2020symbolic, toro2018teaching, camacho2019ltl}) have investigated performing reinforcement learning subject to various types of \gls{ltl} and symbolic constraints.
Work has also investigated adding constraints to demonstrated skills (e.g. \cite{innes2020elaborating, rana2018towards}).

These results are complementary to this paper--here we use the framework in~\cite{innes2020elaborating} to create skills that satisfy our suggested \gls{ltl} constraints.
One can substitute other techniques for the physical repair.

\subsection{Relation to Task and Motion Planning}
Related to this work is the field of \gls{tamp} (for a review see~\cite{caelan2021integrated}).
In \gls{tamp}, actions or skills are often defined symbolically using a STRIPS-style framework \cite{fikes1971strips}, where each skill has a set of preconditions and effects.
Skills in this work also have precondition and postconditions, but also contain information on what occurs during the execution of the skill at the symbolic level.
In this work, tasks are specified as \gls{ltl} formulas that may include sequencing, conditions, and reactions to environment events, while tasks in \gls{tamp} are often to achieve a goal state or set.
Given a task, \gls{tamp} finds a solution by typically combining symbolic search and geometric motion planners.
In contrast, our approach assumes a set of skills that can be composed is given, and we check before execution if the skills can solve the task, and if not, perform repair.
In general, \gls{tamp} is better suited for tasks that require motion and manipulation in highly cluttered environments (e.g. \cite{srivastava2014combined, moll2017randomized, lee2021tree, dantam2018incremental}), while the approach proposed here is better suited for tasks that can be performed with a small curated set of skills but that include complex, temporally-extended goals, reactions to environment events, and conditional safety requirements.

\section{Preliminaries}
In this section we provide background information regarding skills, their abstraction, the logic we use to capture specifications (\gls{ltl}), and techniques for modifying skills.  
\begin{example}[Nine squares]\label{exm:9s}
Consider a 2D environment where the state (x,y) is abstracted by nine squares, as shown in Fig.~\ref{fig:environment_and_symbols}. 
In the following, we will use this simple example to  illustrate different skills, tasks, and the repair process.  
\end{example}

\subsection{Skills and Skill Abstraction}\label{sec:skills_prelim}

\begin{figure}
	\centering
	\includegraphics[width=1\columnwidth]{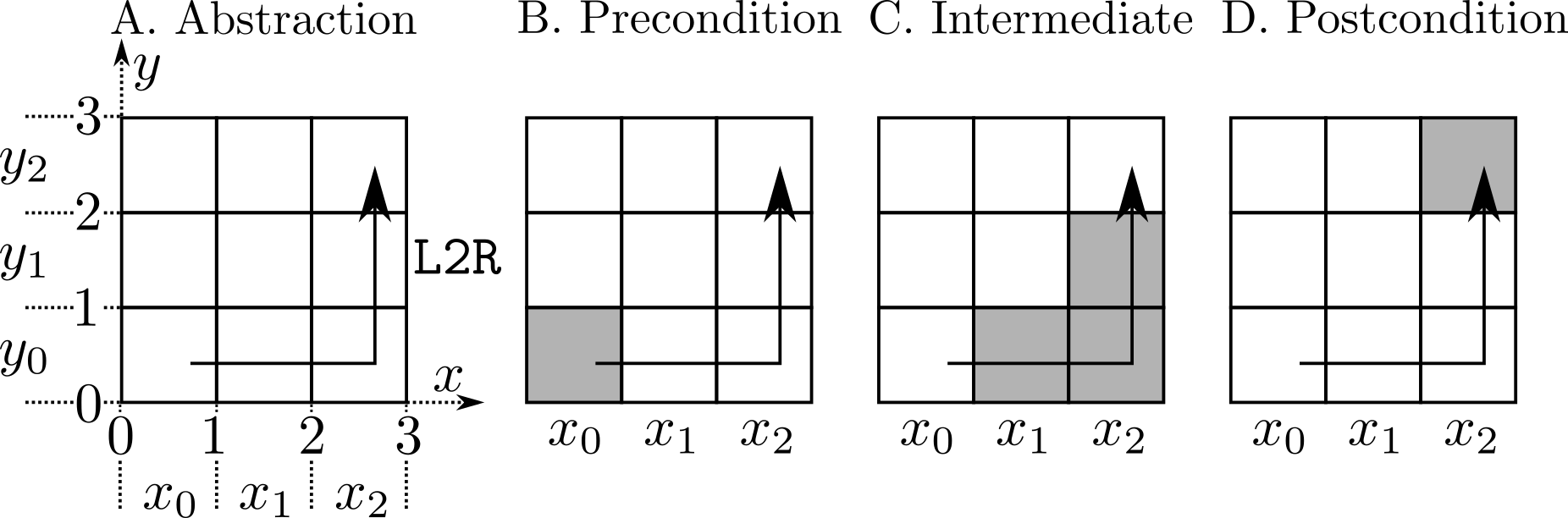}
	\caption{Skill \texttt{L2R} and abstraction.
	A) The $x$ dimension is abstracted into three symbols $x_0$, $x_1$, and $x_2$.
	The $y$ dimension is similarly abstracted. 
	B) The precondition is shown shaded in gray.
	C) The intermediate states are shown shaded in gray.
	D) The postcondition is shown in gray.}
	\label{fig:environment_and_symbols}
\end{figure}

The robot operates in a continuous state space $(x_1,\ldots,x_n)\in X\subseteq \mathcal{R}^n$ with a given set of skills $\skills$.
A $\skill \in \skills$ is associated with a controller that causes the robot to change the physical state of the world.
A skill can only be applied from a specific set of states (the precondition) and results in a set of states (the postcondition).
While a skill is being executed, it may pass through an intermediate set of physical states that are neither the precondition nor the postcondition.
We assume that a skill is an atomic action, that is, once it starts executing, it does not stop until the state is in the postcondition.
In Example~\ref{exm:9s}, seen in Fig.~\ref{fig:environment_and_symbols}, skill \texttt{L2R}'s  precondition is $0\leq x<1$ and $0\leq y < 1$, and its postcondition is $2 \leq x < 3$ and $2 \leq y < 3$.

To enable symbolic reasoning and specifying high-level behaviors, we use a discrete abstraction--a set of Boolean propositions--to describe the state of the world that is then used to define tasks and perform symbolic repair. 
We use the grounding of the abstractions--the physical meaning of each proposition--for the physical repair.
We define $\inpsyms$ as the set of propositions that capture the state of the world. 
Each proposition $\var \in \inpsyms$ grounds to a subset of the state space via the grounding operator $\grounding$ (i.e. $\grounding(\var) \subseteq X$).

We define a \textit{symbolic world state}, $\sigma \subseteq \inpsyms$ as a set of propositions, with the corresponding grounding $\grounding(\statevar) = \bigcap_{\var \in \statevar} \grounding(\var) \bigcap_{\var \in \inpsyms \setminus \statevar} (X \setminus \grounding(\var))$.
In Example~\ref{exm:9s}, $\inpsyms = \{x_0, x_1, x_2, y_0, y_1, y_2\}$ and $\statevar=\{x_0, y_0\}$ grounds to $\grounding(\{x_0, y_0\}) = [0, 1) \times [0, 1)$.

The \textit{initial precondition}, the set of symbolic world states from which a $\skill \in \skills$ can be executed, is given by $\dom^\textrm{pre-init}_\skill \subseteq 2^{\inpsyms}$.
The \textit{final postcondition}, the set of symbolic world states a $\skill \in \skills$ will result in, is given by $\dom^\textrm{post-final}_\skill \subseteq 2^{\inpsyms}$. 
The propositions $\inpsyms$, and the sets $\dom^\textrm{pre-init}_\skill$, and $\dom^\textrm{post-final}_\skill$ can all be automatically created (using, e.g., \cite{konidaris2018skills}); in this work, we assume they are given.

In Example~\ref{exm:9s}, $\dom^\textrm{pre-init}_\texttt{L2R} = \{\{x_0,\allowbreak y_0\}\allowbreak\}$ and $\dom^\textrm{post-final}_\texttt{L2R} = \{\allowbreak\{x_2,\allowbreak y_2\}\}$.
In general, the initial preconditions and final postconditions can consist of multiple symbolic world states.
For example, there could be a skill with a nondeterministic outcome  $\dom_\skill^\textrm{post-final} = \{\{x_2,y_0\},\{x_1,y_1\}\}$, as in following examples.

\subsection{Linear Temporal Logic}\label{sec:ltl}
Formulas in \gls{ltl} are written over a set of atomic propositions, $\vars$, where $\var \in \vars$ is a Boolean proposition.
The syntax of \gls{ltl} is given by the following grammar:
\begin{equation*}
    \varphi \coloneqq \pi \setor \lnot \varphi \setor \varphi_1 \vee \varphi_2 \setor \bigcirc \varphi \setor \varphi_1\ \mathcal{U}\ \varphi_2
\end{equation*}
where $\varphi$ is an \gls{ltl} formula, $\lnot$ is negation, $\vee$ is disjunction, $\bigcirc$ is the next operator, and $\mathcal{U}$ is the until operator.
We define $\true \equiv \varphi \vee \lnot \varphi$ and $\false \equiv \lnot \true$.
Given these operators, one can derive conjunction ($\varphi_1 \wedge \varphi_2 \equiv \lnot(\lnot\varphi_1 \vee \lnot\varphi_2)$), implication ($\varphi_1 \rightarrow \varphi_2 \equiv \lnot\varphi_1 \vee \varphi_2$), equivalence ($\varphi_1 \leftrightarrow \varphi_2 \equiv (\varphi_1 \rightarrow \varphi_2) \wedge (\varphi_2 \rightarrow \varphi_1)$), eventually ($\lozenge \varphi \equiv \true\ \mathcal{U}\ \varphi$), and always ($\square \varphi \equiv \lnot \lozenge \lnot \varphi$).

We use prime ($'$) to denote propositions at the next time step (i.e. $\bigcirc \var$) with $\vars' = \{\var' \setor \var \in \vars\}$.

The semantics of an \gls{ltl} formula $\varphi$ are defined over infinite words $\omega = \statevar_1 \statevar_2 \ldots$ where $\statevar_i \subseteq \vars$ is the set of propositions that are True in step $i$~\cite{pnueli1977temporal}.
We denote that a word $\omega$ satisfies a formula $\varphi$ at step $i$ by $\omega, i \models \varphi$.
Intuitively, $\omega, i \models \bigcirc \varphi$ if $\varphi$ is $\true$ at step $i+1$, $\omega, i \models \lozenge \varphi$ if $\varphi$ is $\true$ at some step after or including $i$, and $\omega, i \models \square \varphi$ if $\varphi$ is $\true$ at all steps after and including $i$.

In this work, we consider the \gls{gr1} fragment of \gls{ltl} \cite{bloem2012synthesis}.
The set of propositions $\vars$ is partitioned into a set of input ($\inp$) and output ($\out$) propositions ($\vars = \inp \cup \out$).
Input propositions $\inp$ correspond to the state of the world, which ground to the physical world the robot can control through its skills ($\inpsyms$) or that are controlled by the user ($\inpuser$).
User controlled propositions could correspond to user inputs such as pressing a button or interacting with a GUI. 
The output propositions $\out$ are controlled by the system (or robot, which we use interchangeably throughout).
In this work, they correspond to the activation of the skills the robot has; each $\skill \in \skills$ has a corresponding $\var_\skill \in \out$.
In Fig.~\ref{fig:nine_squares_example_and_constraints}A, $\inp = \{x_0,\allowbreak x_1,\allowbreak x_2,\allowbreak y_0,\allowbreak y_1,\allowbreak y_2\}$ and $\out = \{\var_\texttt{L2R}, \allowbreak \var_\texttt{R2L}\}$.

In \gls{gr1}, specifications take the form~\cite{bloem2012synthesis}:
\begin{equation}
    \varphi = \varphi_e \rightarrow \varphi_s
\end{equation}
where $\envspec = \envinit \wedge \envsafety \wedge \envlive$ contains assumptions about the environment behavior and $\sysspec = \sysinit \wedge \syssafety \wedge \syslive$ contains requirements on the system's behavior and: 
\begin{flushitemize}
\item $\envinit$ and  $\sysinit$ are Boolean formulas over $\inp$ and $\inp \cup \out$, respectively, characterizing the initial states.
\item $\varphi_{e}^t$ and $\varphi_{s}^t$ are safety constraints of the form $\bigwedge_i \square \psi_i$ where $\psi_i$ are over $v$ and $\bigcirc u$.
For $\varphi_{e}^t$, $v \in \inp \cup \out$ and $u \in \inp$.
For $\varphi_{s}^t$, $v, u \in \inp \cup \out$.
\item $\envlive$ and $\syslive$ are the fairness assumptions and liveness requirements, respectively, that capture events that should occur repeatedly. 
Here, $\envlive = \bigwedge_{i=1}^{m} \square \lozenge J^e_i$ and $\syslive = \bigwedge_{j=1}^{n} \square \lozenge J^s_j$ where  $J^e_i$ and $J^s_j$ are Boolean formulas over $\inp \cup \out$.
\end{flushitemize}

\subsection{Notation: Symbolic States and Corresponding Formulas}
Given the discrete abstraction of the world and the skills, we define a \textit{symbolic state} $\sigma$ as a set of propositions that are $\true$. 
We use $\Sigma_{\alpha} = 2^{\alpha}$ to denote a set of symbolic states and $\statevar_{\alpha} \in \Sigma_{\alpha}$ to denote a single symbolic state where $\alpha\in\{\inp, \inp', \out, \out', \vars, \vars', \inpsyms\}$. 
During the skill encoding process and for feedback from the physical implementation, we only use \textit{symbolic world states} $\sigma \in 2^{\inpsyms}$.
During the repair process, we consider all symbolic states.

To represent a symbolic world state, we use both $\statevar$ and its corresponding Boolean formula $\varphi_\statevar = \bigwedge_{\var \in \statevar} \var \bigwedge_{\var \in \inpsyms \setminus \statevar} \lnot \var$. 
The formula evaluates to $\true$ when the world is in symbolic state $\sigma$. 
We use both representations to simplify the descriptions in the paper: we use the formulas $\varphi_\statevar$ to encode skills in \gls{gr1} and the feedback between the symbolic repair and physical feasibility. 
We use $\statevar$ when describing the repair process.

\subsection{Synthesis}\label{sec:synthesis}

\begin{figure*}
    \centering
    \includegraphics[width=.8\textwidth]{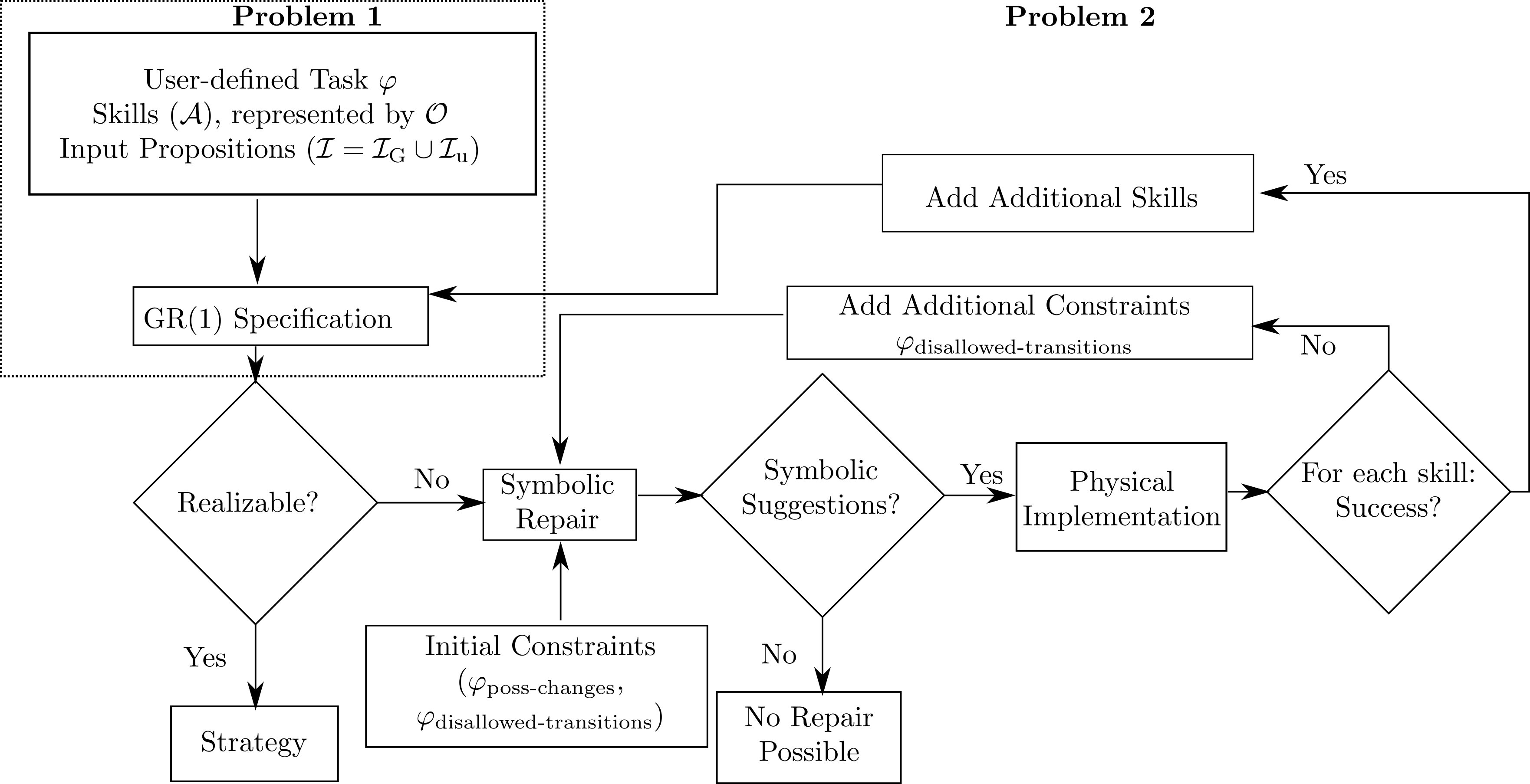}
    \caption{Overview of our approach to task encoding and repair. 
    }
    \label{fig:flowchart}
\end{figure*}

We use GR(1) synthesis~\cite{bloem2012synthesis} to find a strategy, a symbolic composition of skills, to accomplish a task.
Synthesis involves solving a two-player game played between the system (robot) and its environment, where the environment plays first and the system reacts to it. 
Synthesis, as done in~\cite{bloem2012synthesis}, is a worst-case approach--the environment is considered to be adversarial and it attempts to keep the system from accomplishing its task.
This ensures that if synthesis returns a strategy, the system is guaranteed to accomplish its task regardless of what the environment does, including changes to user controlled propositions.

Following~\cite{bloem2012synthesis}, we define a \emph{game structure} as $\game = (\vars, \inp, \out, \initstates, \tenv, \tsys, \obj)$ %, 
where $\vars$ is the set of propositions and $\inp$ and $\out$ are defined as in Section~\ref{sec:ltl}.
We define $\initstates$ as the set of states that satisfy $\envinit \wedge \sysinit$.
We define $\tenv \subseteq \allstates \times \dom_{\inp'}$ as the set of current states and next input states satisfying $\envsafety$ and 
$\tsys \subseteq \allstates \times \dom_{\vars'}$ as the set of current and next states satisfying $\syssafety$.
Note that in \cite{bloem2012synthesis}, $\tenv$ and $\tsys$ are defined as logical formulas; here we define them as sets of sets of propositions.
The winning condition is given by $\obj = \envlive \rightarrow \syslive$.

Given a game structure $\game$, the \emph{realizability} problem is to decide if the game is winning for the system; either (a) for every environment action, the system is able to achieve $\varphi_s$ or (b) the system is able to falsify $\varphi_e$.
To determine if a specification is realizable, we find, via a fixed point computation, all the states $Z \subseteq \allstates$ from which the system is able to win \cite{bloem2012synthesis}.
We iterate through every system liveness guarantee, $J^s_j$, and determine the set of states the system can always either transition to the next liveness goal from or falsify $\varphi_e$.
The \emph{synthesis} problem is to compute a strategy for the system to make the specification realizable~\cite{bloem2012synthesis}.

\subsection{Physically Implementing Skills}\label{physical_skills_prelim}

There are many ways to physically implement skills for a robot (e.g. \cite{li2017reinforcement, innes2020elaborating, rana2018towards}).
In this work, we use \cite{innes2020elaborating} for the physical feasibility checking and implementation of modified skills.
There, given a set of demonstrations (trajectories), the authors transform \gls{ltl} constraints into a loss function and use that to generate new trajectories that remain close to the original demonstrated trajectories while satisfying the \gls{ltl} constraints. 
Using the \gls{ltl} loss and another loss term that captures how close a new trajectory is to the original trajectory, \cite{innes2020elaborating} can tradeoff between satisfying the \gls{ltl} specification and remaining close to the original trajectory.

Specifically, \cite{innes2020elaborating} models a skill as a \gls{dmp} \cite{schaal2005learning}, which is a series of differential equations that use radial basis functions to express non-linear  trajectories.
For a given skill, the start and end point are supplied to the \gls{dmp} and the full discretized trajectory of the skill in $X$ is given.
\cite{innes2020elaborating} train a feed-forward neural network to generate the weights associated with the radial basis functions using the \gls{ltl} loss function and use these \gls{dmp}s to generate trajectories, or more precisely, a series of waypoints, that satisfy the provided \gls{ltl} constraints.

\section{Problem Statement}

Given a robot with a set of skills ($\skills$) represented using output propositions ($\out$), a set of input propositions ($\inp = \inpsyms \cup \inpuser$), and a user provided (reactive) task $\varphi$, our goal is to find a set of skills and a composition strategy such that the robot can physically accomplish the user provided task.

We split this into two problems, (1) the encoding of the skills in a \gls{gr1} formula and synthesis, and (2) the repair process (symbolic and physical) when a task is unrealizable, given the skills.
For the second problem, we allow the user to optionally provide constraints on the repair of the task, as discussed in the following.

\textbf{Problem 1:} Given a robot with a set of skills $\skills$, encode the skills in a \gls{gr1} formula $\varphi$ such that safety constraints can be enforced on the intermediate states of skills and livenesses can be achieved during the execution of skills. 

\textbf{Problem 2:} Given a \gls{gr1} formula $\varphi$ encoding the skills $\skills$ of a robot and the task, the controllers implementing the skills, the proposition groundings ($\grounding$), constraints on which propositions can change, and physically impossible transitions, find physically-implementable skills that result in the robot being able to achieve $\varphi$. 

\section{Approach}
We present our approach to solving Problems 1 and 2 in Fig.~\ref{fig:flowchart}. 
First, we encode the task and skills as a \gls{gr1} formula.
If the specification is realizable, we create a strategy that the robot can execute.
If it is not realizable, we perform symbolic repair and generate symbolic suggestions for skills that respect both user provided and automatically generated constraints.
Next, we attempt to physically implement the suggestions.
If we are successful in physically implementing a skill, we encode it in the specification.
If not, we add a constraint that precludes the physically infeasible suggestion.
We then synthesize again to check whether the task is realizable, and if it is not, we repeat the repair.

Two things to note: (i) the repair is performed offline; after the system is repaired and a strategy is generated, we can guarantee the robot will accomplish the desired task, and (ii) the repair process only adds additional skills, it does not remove any of the original skills. 
If a skill should be removed, the user can forbid it from being executed.

\section{Encoding Skills as GR(1) formulas}

\begin{figure*}[b]
    \addtocounter{equation}{1}
\begin{equation}\label{eq:sys_trans_choose}
    \begin{split}
        &\syssafetychoose = \bigwedge_{\var_\skill \in \out} \square \lnot \left( \bigvee_{\statevar_\textrm{pre-init} \in \dom^\textrm{pre-init}_\skill} \bigcirc \varphi_{\statevar_\textrm{pre-init}} \bigvee_{\statevar_\textrm{post} \in \dom_\skill^\textrm{only-intermediate}}  \bigvee_{\statevar_\textrm{pre} \in \dom_{\skill,\statevar_\textrm{post}}^\textrm{pre}} \left(\varphi_{\statevar_\textrm{pre}} \wedge \pi_\skill \wedge \bigcirc \varphi_{\statevar_\textrm{post}} \right) \right) \rightarrow \lnot \bigcirc \pi_\skill
    \end{split}
\end{equation}
\begin{equation}\label{eq:sys_trans_continue}
    \begin{split}
        &\syssafetycontinue = \bigwedge_{\var_\skill \in \out} \bigwedge_{\statevar_\textrm{post} \in \dom_\skill^\textrm{only-intermediate}}  \bigwedge_{\statevar_\textrm{pre} \in \dom_{\skill,\statevar_\textrm{post}}^\textrm{pre}} \square \left(\varphi_{\statevar_\textrm{pre}} \wedge \var_\skill \wedge \bigcirc \varphi_{\statevar_\textrm{post}} \right) \rightarrow \bigcirc \var_\skill
    \end{split}
\end{equation}
\end{figure*}
\begin{figure}
	\centering
	\includegraphics[width=1\columnwidth]{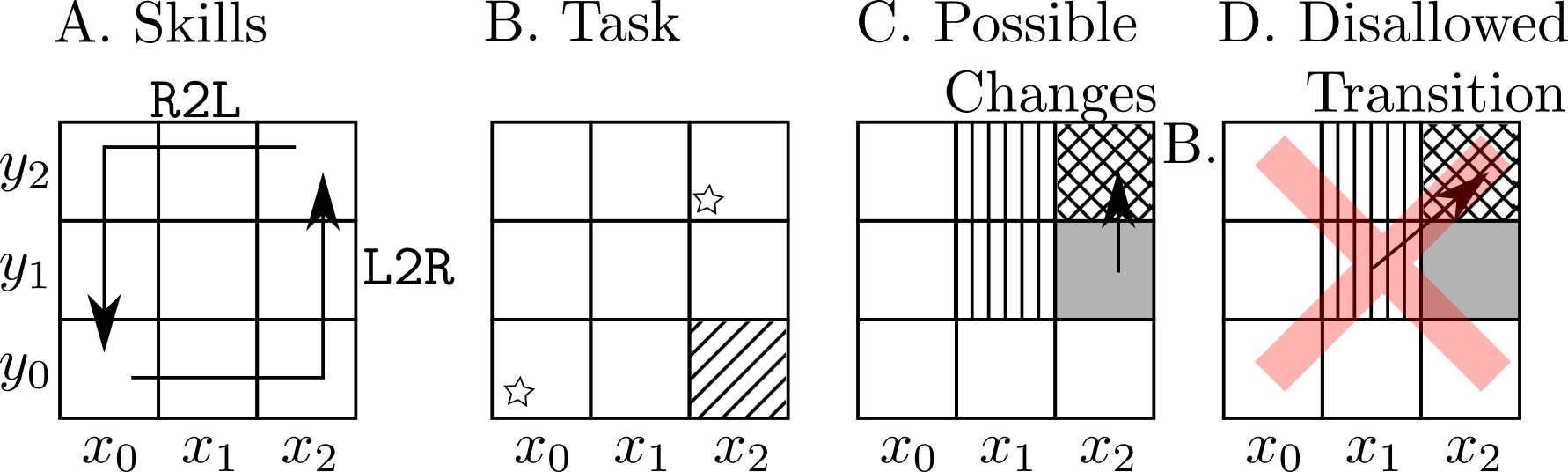}
	\caption{
	A-B) Example~\ref{exm:9s}: Nine Squares. 
	A) The robot is given two skills, \texttt{L2R} and \texttt{R2L}. 
	There are 6 propositions, 3 each for the $x$ and $y$ dimensions. 
	B) The task is to always eventually visit the top right corner and bottom left corner, i.e. $\syslive = \square \lozenge (x_2 \wedge y_2) \wedge \square \lozenge (x_0 \wedge y_0)$, (shown with the stars) while avoiding the bottom right corner, i.e. $\syssafetyuser = \square \lnot (x_2 \wedge y_0) \wedge \square \lnot \bigcirc (x_2 \wedge y_0)$.
	Initially the task is unrealizable because the skill $\texttt{L2R}$ must go through $x_2 \wedge y_0$ in order to reach $x_2 \wedge y_2$.
	C-D) Repair constraints. 
	C) The possible changes to the precondition $\{x_2,y_1\}$ for the skill shown by the black arrow subject to $\changecons = \square (x_2 \wedge \lnot \bigcirc (x_0 \vee x_2)) \wedge \square (y_1 \wedge \lnot \bigcirc y_0)$.
	The current precondition $\{x_2,y_1\}$ is shown as the shaded square, the current postcondition $\{x_2,y_2\}$ is shown as the double hashed square.
	The possible modified preconditions are shown by the vertical hashes.
	D) A modification that is not physically implementable going from the precondition $\{x_1,y_1\}$ to the postcondition $\{x_2,y_2\}$ is disallowed by $\tnotallowedrepair = \square \lnot (x_1 \wedge y_1 \wedge \var_\texttt{skill} \wedge \bigcirc x_2 \wedge \bigcirc y_2)$.}
	\label{fig:nine_squares_example_and_constraints}
\end{figure}
In this work we are given a set of skills the robot has and we assume all related propositions and their groundings are given as well; either determined automatically (using, e.g., \cite{konidaris2018skills}) or provided by the user.  
Prior work \cite{pacheck2019automatic} has shown how skills can be automatically encoded into \gls{gr1} formulas and used to synthesize robot behaviors. 
In contrast to that work, here, in addition to capturing the preconditions and postconditions on a skill, we also explicitly consider the state of the world \textit{during} the execution of a skill. 
This captures the non-instantaneous nature of robot control and allows the user to receive safety guarantees on the full execution of the robot.
Furthermore, instead of being required to satisfy a liveness guarantee at the end of a skill, it can be satisfied during the skill execution. 

We define $\dom_\skill^\textrm{unique} \subseteq 2^{\inpsyms}$ as the set of unique symbolic world states visited during the execution of $\skill$.
From a given symbolic world state, $\statevar$, we define all possible next states when executing $\skill$ as $\dom_{\skill,\statevar}^\textrm{post} \subseteq 2^{\inpsyms}$.
Similarly, we define all the possible symbolic world states from which $\statevar$ could be reached by executing $\skill$ as $\dom_{\skill,\statevar}^\textrm{pre} \subseteq 2^{\inpsyms}$.
For convenience, we also define the set of states that are only intermediate states, i.e. not initial preconditions or final postconditions, as $\dom_\skill^\textrm{only-intermediate} = \dom_\skill^\textrm{unique} \setminus (\dom_\skill^\textrm{pre-init} \cup \dom_\skill^\textrm{post-final})$.
In Fig.~\ref{fig:environment_and_symbols}, for example, $\dom^\textrm{unique}_\texttt{L2R} = \{\{x_0, y_0\}, \allowbreak \{x_1, y_0\}, \allowbreak \{x_2, y_0\}, \allowbreak \{x_2, y_1\}, \allowbreak \{x_2, y_2\}\}$, $\dom_{\texttt{L2R},\{x_0,y_0\}}^\textrm{post} = \{\{x_1, \allowbreak y_0\}\}$, and $\dom_{\texttt{L2R},\{x_1,y_0\}}^\textrm{pre} = \{\{x_0, \allowbreak y_0\}\}$.

In this work, we do not assume these intermediate states are given--rather we determine  $\dom_{\skill,\statevar}^{\textrm{post}}$ and $\dom_{\skill, \statevar}^{\textrm{pre}}$ by performing multiple physical executions of $\skill$ and monitoring the result.

We automatically encode the skills in the \gls{gr1} formula as part of $\envsafety$ and $\syssafety$.
For the purposes of repair, we define the environment safety formula $\envsafety = \envsafetynothard \wedge \envsafetyhard$ where $\envsafetynothard$ are safety formulas that the repair process may modify; they capture the (possibly non-deterministic) post-conditions, or outcomes, of skills.
We encode safety formulas that may not change, for example mutual exclusion of propositions due to their groundings, in $\envsafetyhard$; if the user provided assumptions about the environment, $\envsafetyharduser$, we add it to $\envsafetyhard$ thereby not allowing the repair to modify those constraints.

The system safety formula, $\syssafety$, encodes constraints on when skills can be executed, based on the pre-conditions.
Similar to $\envsafety$, we define $\syssafety = \syssafetynothard \wedge \syssafetyhard$, where $\syssafetynothard$ are formulas the repair process may modify, and $\syssafetyhard$ are formulas that remain unchanged. 
We consider user provided safeties, $\syssafetyuser$, as part of $\syssafetyhard$. 

\paragraph{Encoding outcomes of skills in $\envsafetynothard$}
We encode the possible postconditions in $\envsafetynothard$ in Eq.~\eqref{eq:env_trans} ; it encodes that for every skill precondition, the resulting state is one of the possible postconditions of that skill.
We show $\envsafetynothard$ for skill \texttt{L2R} in Eq.~\eqref{eq:tenv_lefttoright} in the Appendix.
\addtocounter{equation}{-3}
\begin{equation}\label{eq:env_trans}
\begin{split}
    \envsafetynothard = \bigwedge_{\var_\skill \in \out} \bigwedge_{\statevar \in \dom_\skill^\textrm{unique} \setminus \dom^\textrm{post-final}_\skill} \square& \bigg( \left( \pi_\textrm{skill} \wedge \varphi_{\statevar}\right) \rightarrow\\
    &\bigcirc \bigvee_{\statevar_\textrm{post} \in \dom_{\skill,\statevar}^\textrm{post}} \varphi_{\statevar_\textrm{post}}\bigg)
    \end{split}
\end{equation}
\addtocounter{equation}{2}

\paragraph{Encoding constraints on skills in $\syssafetynothard$}
We encode preconditions and requirements stemming from intermediate states in $\syssafetynothard = \syssafetychoose \wedge \syssafetycontinue$, as shown in  Eqs.~\eqref{eq:sys_trans_choose} and \eqref{eq:sys_trans_continue}.
The formula $\syssafetychoose$ restricts when a skill can be activated; it prevents the system from choosing a skill when (i) none of the initial preconditions are satisfied and (ii) the skill is not in the middle of being executed (i.e. the system is not in an intermediate state of the skill).
This therefore allows the system to choose a skill only when a precondition is satisfied, but does not require the system to do so.
The formula $\syssafetycontinue$ requires the system to continue executing a skill from intermediate states it reaches during execution, i.e. a skill cannot be aborted in the middle of the execution.
We provide an example of $\syssafetynothard$ for skill $\textrm{L2R}$ in Eq.~\eqref{eq:tsys_lefttoright} in the Appendix.

\paragraph{Hard constraints}
As described above, certain safety constraints may not be changed by the repair process. We encode these in $\syssafetyhard$ and $\envsafetyhard$.
In $\syssafetyhard$, we require that at most one skill is active at any given time.
In $\envsafetyhard$, we encode mutual exclusion of propositions that are grounded to disjoint sets (i.e. $\grounding(\var_1) \cap \grounding(\var_2) = \emptyset$).
For example, for the abstraction in Fig.~\ref{fig:nine_squares_example_and_constraints}, $\square \lnot(x_0 \wedge x_1)$ is part of $\envsafetyhard$. 
We also require all propositions in $\inpsyms$ to remain unchanged when no skill is active, i.e. $\square ((\bigwedge_{\var_\mathcal{O} \in \out} \lnot \var_\mathcal{O}) \rightarrow (\bigwedge_{\var_\mathcal{I} \in \inpsyms} \var_\mathcal{I} \leftrightarrow \bigcirc \var_\mathcal{I}))$. 

\section{Symbolic Repair}\label{sec:symbolic_algorithms}

When a specification is unrealizable, we perform symbolic repair, which returns symbolic suggestions for skill modifications that would allow the task to be completed.
Similar to \cite{pacheck2020finding}, we modify either the preconditions or the postconditions of skills;  we either relax preconditions to make skills applicable from more states, or modify postconditions which can reduce a skill's nondeterminism (and thereby not allow the skill to enter unwanted states). 
However, in contrast to \cite{pacheck2020finding}, here we (i) allow the repair process to modify the intermediate states, (ii) always modify existing skills (i.e. we do not create new skills), and (iii) allow the user and/or the physical repair to enforce constraints on the symbolic repair. 

We present the algorithm for modifying the preconditions first (Sec.~\ref{sec:modify_pre}) and then the algorithm for modifying the postconditions (Sec.~\ref{sec:modify_post}).
However, when performing repair, we also generate suggestions by modifying the postconditions first and then the preconditions, as the resulting suggestions may be different.

We use Example~\ref{exm:9s} to illustrate the repair process.
Here, we are given two skills (\texttt{L2R} and \texttt{R2L}) as shown in Fig.~\ref{fig:nine_squares_example_and_constraints}A.
The user-specified task is ``repeatedly visit the upper right and lower left corners ($\syslive = \square \lozenge (x_2 \wedge y_2) \wedge \square \lozenge (x_0 \wedge y_0)$) while always avoiding the lower right corner ($\syssafetyuser = \square \lnot (x_2 \wedge y_0) \wedge \square \lnot \bigcirc (x_2 \wedge y_0)$) starting from the lower left corner ($\sysinit = x_0 \wedge y_0$) as shown in Fig.~\ref{fig:nine_squares_example_and_constraints}B.

\subsection{Constraints}\label{sec:constraints}

In this section we discuss constraints we impose on the repair process.
We describe constraints on how input propositions can change  (Sec.~\ref{sec:input_constraints}) and on modifications that are known to be physically impossible (Sec.~\ref{sec:not_allowed_repair}).
\begin{algorithm*}[h]
\KwIn{$\tenv$, $\tenvhard$, $\tsysnothard$, $\tsyshard$, $Z$, $\inpuser$}
\KwOut{$\tenvnew$, $\tsysnewmutable$}
\BlankLine
\tcp{States where the system is guaranteed to reach $Z$ for any environment choice}

$\tcanwinandencoded = \textrm{systemAlwaysWinsAndSkillsAreReachable}(\tsysnothard, \tenv, Z)$\;\label{line:simple_pre_canwinandencoded}

\tcp{Find allowed changes to preconditions}
$\tallowablechanges = \textrm{findAllowableChanges}(\tcanwinandencoded, \tchangeconstraints, \tnotallowedrepair, \tsyshard, \tenv)$\;\label{line:simple_pre_allowablechanges}

\tcp{Randomly select the possible change to add}
\eIf{$\tallowablechanges \neq \emptyset$}{
$\tselectedchange = \textrm{randomlySelectChange}(\tallowablechanges, \tsysnothard)$\label{line:simple_pre_selectedchange}\;
$\tselectedchange = \textrm{removeDependenceOnUserPropositions}(\tselectedchange, \inpuser)$\;\label{line:simple_line_pre_alluservariables}
}
{
\Return $\tenv$, $\tsysnothard$\;
}
$\tnewfullskill = \textrm{replacePreconditions}(\tselectedchange)$\;\label{line:simple_pre_newfullskill}

\tcp{Add transitions to $\tsysnothard$}
$\tnewpreprecondition = \textrm{addPrePrecondition}(\tselectedchange, \tsysnothard)$\;\label{line:simple_pre_newprepreconditions}

\tcp{Determine if this is the initial precondition}
\If{$\textrm{isInitialPrecondition}(\tnewpreprecondition)$}{\label{line:simple_pre_modinitprecheck}
$\tnewpreprecondition =\textrm{disallowExecutionFromOtherIntermediateStates}(\tsysnothard, \tnewpreprecondition)$\;\label{line:simple_pre_newprepreconditions2}
}
$\tsysnewmutable = \tsysnothard \cup \tnewpreprecondition \cup \tnewfullskill$\;\label{line:simple_pre_sysnothardnew}

\tcp{Add transitions to $\tenv$}

$\tenvnew = \textrm{createNewEnvTransition}(\tenv, \tenvhard, \tnewfullskill, \tselectedchange, \tsysnothard)$\;\label{line:simple_pre_envnew2}
\Return $\tenvnew$, $\tsysnewmutable$
\caption{\textbf{modify\_preconditions}}
\label{alg:modifyPre_simple}
\end{algorithm*}

\subsubsection{Constraints on changes to $\inpsyms$}\label{sec:input_constraints}
We allow the user to restrict, if they choose to, how symbolic world states $\sigma \in 2^{\inpsyms}$ (i.e. the truth values of the world input propositions $\inpsyms$) are allowed to change in the repair process. 

We specify the allowed modifications as $\changecons = \bigwedge_i \square \varphi_i$ where $\varphi_i$ is a Boolean formula over $\inp \cup \bigcirc \inp$. 
This formula is only used in the repair process and is not part of the task when synthesizing. 
We define the set $\tchangecons \subseteq \inpstates \times \inpstatesprime$ as pairs of input states that satisfy $\changecons$.
During the repair process, the preconditions (or postconditions) of a skill can be modified from $\inpstate$ to $\inpstate'$ only if $(\inpstate, \inpstate') \in \tchangecons$.

We show an example of $\changecons$ in Fig.~\ref{fig:nine_squares_example_and_constraints}C.
Consider a skill that goes from $\{x_2,y_1\}$ to $\{x_2, y_2\}$.
If $\changecons = \square (x_2 \wedge \lnot \bigcirc (x_0 \vee x_2)) \wedge \square (y_1 \wedge \lnot \bigcirc y_0)$, we are not allowed to change $x_2$ to either $x_0$ or $x_2$ and are also not allowed to change $y_1$ to $y_0$.
This only allows us to modify the preconditions such that a candidate modified skill goes from $\{x_1,y_1\}$ to $\{x_2,y_2\}$ or from $\{x_1,y_2\}$ to $\{x_2,y_2\}$ as shown in Fig.~\ref{fig:nine_squares_example_and_constraints}C.

\subsubsection{Disallowed Transitions Due to Physical Infeasibility}\label{sec:not_allowed_repair}

We place constraints on which transitions can be suggested for a given skill; 
we primarily use these constraints to automatically incorporate feedback from the physical implementation process where we find transitions that are not physically possible, although these constraints can be given by the user as well.
We write these constraints as $\notallowedrepair = \bigwedge_i \square \lnot \varphi_i$ where $\varphi_i$ is a Boolean formula over $\vars \cup \bigcirc \inp$.
We create the set $\tnotallowedrepair \subseteq \allstates \times \inpstatesprime$ which contains pairs of states that satisfy $\notallowedrepair$.
As with $\tchangecons$, $\tnotallowedrepair$ is only used in the repair process, not the synthesis process.

For example, if $\notallowedrepair = \square \lnot (x_1 \wedge y_1 \wedge \var_\texttt{skill} \wedge \bigcirc x_2 \wedge \bigcirc y_2)$, the repair process cannot suggest a modified skill going from $\{x_1,y_1\}$ to $\{x_2,y_2\}$, as shown in Fig.~\ref{fig:nine_squares_example_and_constraints}D.

\begin{figure}
	\centering
	\includegraphics[width=.75\columnwidth]{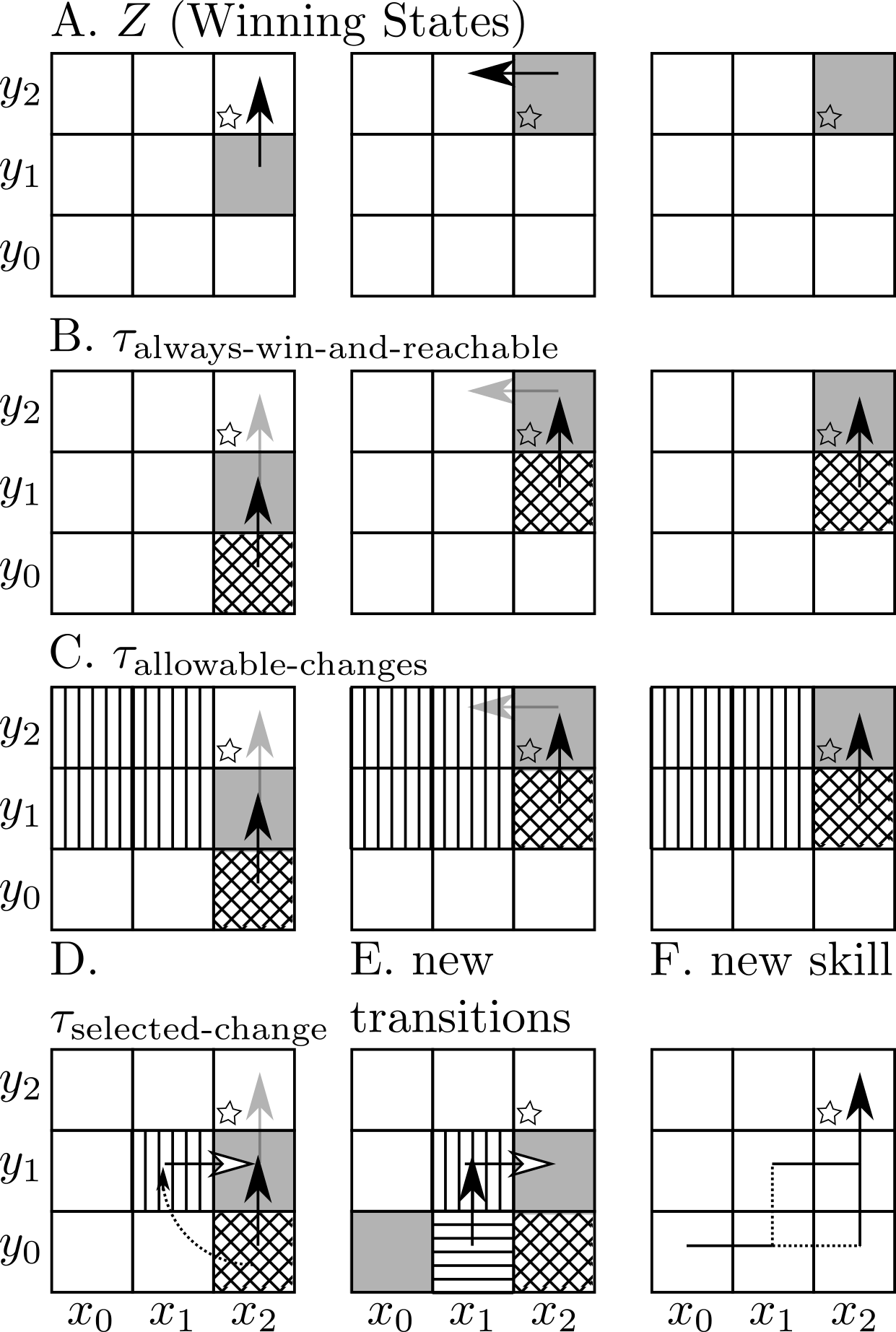}
	\caption{Overview of modifying the preconditions for Example~\ref{exm:9s}.
	(A) The winning states $Z$ when the system is trying to satisfy the liveness $\square \lozenge (x_2 \wedge y_2)$. 
	The input states are the shaded squares and the arrows are the skill being applied.
	(B) Transitions that can always reach the winning states in (A) and whose preconditions are reachable. 
	The double hashed squares are the precondition, the dark arrow the current skill, the shaded squares the postcondition, and the light arrows the next skill. 
	(C) The allowable changes to the preconditions, shown as vertical hashes. 
	The original preconditions are allowed to be modified to any of the squares with vertical hashes.
	We only restrict that the changes to the input propositions must respect the mutual exclusion of symbols in the $x$ and $y$ dimension in $\tchangecons$. 
	The requirement that a state not already be in the skill does not allow possible preconditions to have states where $y_0$ or $x_2$ are $\true$.
	(D) One selected precondition modification, along with the postcondition it goes to.
	The vertical hashed square is the new precondition, the arrow with the white head is the new portion of the skill, the dark arrow the current skill, the shaded square the postcondition, and the light arrow the next transition in the skill.
	The original precondition is shown with the double hashes.
	(E) The added transitions.
	The transitions associated with the white and black headed arrows are added to the system and environment transitions.
	The black headed arrow is the transition that allows the white headed arrow to be reached.
	The precondition to the original precondition (double hashes) is shown with horizontal hashes. 
	(F) The complete new skill.
	The skill now has a non-deterministic transition as shown by the dashed lines from $\{x_1, y_0\}$ to either $\{x_2, y_0\}$ or $\{x_1, y_1\}$.
	}
	\label{fig:nine_squares_modify_pre}
\end{figure}

\subsection{Modify Preconditions}\label{sec:modify_pre}

Given a skill, the physical meaning of modifying its preconditions is allowing it to be executed from a different set of initial states, with the goal of making the task realizable.
Given the current environment transitions ($\tenv$), the current hard environment transitions ($\tenvhard$), the set of changeable system transitions ($\tsysnothard$), the set of hard system transitions ($\tsyshard$), and the set of winning states ($Z$), we find skills that always reach $Z$.
We then modify the preconditions of those ``winning" skills, subject to $\tchangecons$ and $\tnotallowedrepair$, to enable the system to execute these skills from more states; we add the modified preconditions to the existing preconditions of that skill in $\tsysnothard$.
We also modify $\tenv$ to allow the new preconditions to reach the appropriate postcondition and if necessary be reached (if it is an intermediate state) by the modified skill.
We present our algorithm for modifying preconditions in Alg.~\ref{alg:modifyPre_simple} and in more detail in Alg.~\ref{alg:modifyPre} in the Appendix.
Line numbers in this section reference Alg.~\ref{alg:modifyPre_simple}.

We find the set $\tcanwinandencoded \subset \allstates \times \allstatesprime$ (Line~\ref{line:simple_pre_canwinandencoded}) such that $\statevar'\in \allstatesprime$ is a winning state (i.e. $\statevar' \in \winning$) and $\statevar\in \allstates$ is reachable, i.e. there exists a transition in $\tenv$ that results in $\statevar$. 

In Example~\ref{exm:9s}, the repair process is triggered when the synthesis process attempts to reach $\{x_2, y_2\}$.
Here, $Z =\{\{x_2,y_1,\var_\texttt{L2R}\},\allowbreak\{x_2,y_2, \var_\texttt{R2L}\},\allowbreak\{x_2, y_2\}\}$ as shown in Fig.~\ref{fig:nine_squares_modify_pre}A.
We find $\tcanwinandencoded = \big\{\{x_2, y_0, \var_\texttt{L2R}, x_2', y_1', \var_\texttt{L2R}'\},\allowbreak\{x_2, y_1, \var_\texttt{L2R}, x_2', y_2', \var_\texttt{R2L}'\},\allowbreak\{x_2, y_1, \var_\texttt{L2R}, x_2', y_2'\}\}$ (Fig.~\ref{fig:nine_squares_modify_pre}B).

We then find allowable changes to the preconditions; we capture allowable changes by specifying current states $\allstates$, next states $\allstatesprime$ and possible new preconditions $\inpstatesdoubleprime$ in $\tallowablechanges \subset \allstates \times \allstatesprime \times \inpstatesdoubleprime$
in Line~\ref{line:simple_pre_allowablechanges}\footnote{In Alg.~\ref{alg:modifyPre}, we use dagger ($^{\dagger}$) to denote possible changes to input variables (in this section the preconditions) during the repair process where $\inp^{\dagger} = \{\var^{\dagger} \setor \var \in \inp \}$ and $\dom_{\inp^{\dagger}} = 2^{\inp^{\dagger}}$.}.
The changes to $\tcanwinandencoded$ are such that the new preconditions are allowed by the change constraints $\tchangecons$, do not violate $\tnotallowedrepair$, and satisfy the hard system constraints $\tsyshard$.
We also require that the new preconditions are not an intermediate state already in the skill.

In Example~\ref{exm:9s}, the allowable changes are shown in Fig.~\ref{fig:nine_squares_modify_pre}C.
The current preconditions, shown with double hashes, of $\{\{x_2, y_0\},\allowbreak\{x_2,y_1\}\}$ can be modified to be any of the states shown with vertical hashes, i.e. $\{\{x_0,y_1\},\allowbreak\{x_0,y_2\},\allowbreak\{x_1,y_1\},\allowbreak\{x_1,y_2\}\}$. 
As opposed to the example in Fig.~\ref{fig:nine_squares_example_and_constraints}C, here we only require that the changes to the input propositions respect the mutual exclusion of propositions in $\tchangecons$ (e.g. $\square \lnot \bigcirc(x_0 \wedge x_1)$).
Requirements that the new preconditions are not already states of the skill additionally disallow states where $y_0$ or $x_2$ are $\true$.

From these possible precondition modifications, we randomly choose a modification in Line~\ref{line:simple_pre_selectedchange}.
Here we assume that the selected modification is $\tselectedchange =\{x_2,\allowbreak y_0,\allowbreak \var_\texttt{L2R}, \allowbreak x_2', \allowbreak y_1', \allowbreak \var_\texttt{L2R}', \allowbreak x_1^\dagger, \allowbreak y_1^\dagger\}$ as shown in Fig.~\ref{fig:nine_squares_modify_pre}D.
Note that $\tselectedchange$ may contain more than one modification. 
For example, if we modified the precondition $\{x_2,y_1\}$, $\tselectedchange$ would include both $\{x_2,\allowbreak y_1, \allowbreak \var_\texttt{L2R}, \allowbreak x_2', \allowbreak y_2', \allowbreak \var_\texttt{R2L}', \allowbreak x_1^{\dagger}, \allowbreak y_1^{\dagger}\}$ and $\{x_2,\allowbreak y_1, \allowbreak \var_\texttt{L2R}, \allowbreak x_2', \allowbreak y_2', \allowbreak x_1^{\dagger}, \allowbreak y_1^{\dagger}\}$ (the difference being the skill or lack thereof in the next state). 

Since we do not allow our modifications to restrict user-controlled propositions, in Line~\ref{line:simple_line_pre_alluservariables} we add all possible combinations of user controlled propositions $\inpuser$, if they exist, to the suggested change.
Essentially, we are creating multiple copies of the modification, one for each unique truth assignment to the user controlled propositions, thus encoding that those propositions can take on any value.

We add the new (i.e. modified) skill ($\tnewfullskill$, Line~\ref{line:simple_pre_newfullskill}) to $\tsysnothard$ and $\tenv$.
In Example~\ref{exm:9s}, $\tnewfullskill = \{x_1,\allowbreak y_1,\allowbreak \var_\texttt{L2R},\allowbreak x_2',\allowbreak y_1',\allowbreak \var_\texttt{L2R}'\}$ (Fig.~\ref{fig:nine_squares_modify_pre}D).

In order to ensure that the new precondition is reachable (otherwise it cannot be executed and thus cannot repair the task), we determine the preconditions of the original precondition and allow the modified precondition to be executed from those states ($\tnewpreprecondition$, Line~\ref{line:simple_pre_newprepreconditions}).
In Example~\ref{exm:9s}, originally $\texttt{L2R}$ goes from $\{x_1,y_0\}$ to $\{x_2,y_0\}$.
For the modified $\texttt{L2R}$, we add $\{x_1,\allowbreak y_0,\allowbreak \var_\texttt{L2R}, \allowbreak x_1', \allowbreak y_1', \allowbreak \var_\texttt{L2R}'\}$ to $\tsysnothard$ thereby allowing the skill to reach $\{x_1,y_1\}$ from $\{x_1,y_0\}$ as shown by the black headed arrow in Fig.~\ref{fig:nine_squares_modify_pre}E. 
Two things to note here: (i) the added transitions might be physically impossible--they are not subject to $\tnotallowedrepair$--such modifications will be automatically rejected by the physical feasibility checking, and (ii) we are adding transitions but not removing any--essentially we are adding non-determinism to the skill.

If the modified initial preconditions, $\tnewpreprecondition$, allows the skill to be executed from the new preconditions no matter the previous skill, when we add $\tnewpreprecondition$ to $\tsysnothard$, this can override the requirement that a skill continue executing until it reaches its final postcondition.
We therefore adjust $\tnewpreprecondition$ when the initial precondition is modified to prevent the strategy from switching skills mid-execution.
For example, if we attempt to modify the initial precondition $\{x_0,y_0\}$ of $\texttt{L2R}$ to $\{x_0,y_1\}$, $\tnewpreprecondition = \bigcup_{\statevar \in \allstates} \statevar \cup \{x_0',\allowbreak y_1',\allowbreak \var_\texttt{L2R}'\}$.
When added to $\tsysnothard$, this would allow the robot to execute skill $\texttt{L2R}$ from $\{x_0,y_1\}$ instead of being required to continue executing skill $\texttt{R2L}$ until $\{x_0, y_0\}$ is reached.
We do not want this, so we remove the offending transitions from $\tnewpreprecondition$ in Line~\ref{line:simple_pre_newprepreconditions2}.

The new mutable system transitions are the original mutable transitions with the new preconditions and the preconditions that allow that precondition to be reached (Line~\ref{line:simple_pre_sysnothardnew}).

For $\tenvnew$, we add the new precondition in $\tnewfullskill$ and enforce that it reaches the original postcondition (Line~\ref{line:simple_pre_envnew2}).
We also add the precondition and postcondition in $\tnewpreprecondition$.
In Example~\ref{exm:9s}, we add $\{x_1,y_0,\var_\texttt{L2R},x_1',y_1'\}$ and $\{x_1,y_1,\var_\texttt{L2R},x_2',y_1'\}$ to $\tenv$ and disallow $\{x_1, y_1, \var_\texttt{L2R}\} \bigcup_{\statevar\in\Sigma_{\inpsyms'}\setminus\{x_2',y_1'\}}\statevar$ in $\tenv$, essentially not allowing transitions from $(x_1,y_1)$ to any state other than $(x_2,y_1)$.
We show the resulting new skill in Fig.~\ref{fig:nine_squares_modify_pre}F.

\subsection{Modify Postconditions}\label{sec:modify_post}

\begin{algorithm}
\KwIn{$\tenv$, $\tsysnothard$, $Z$, $\inpuser$} 
\KwOut{$\tenvnew$}

\tcp{Find states that do not lead to winning states}
$\tfullskills = \textrm{findReachableSkills}(\tsysnothard, \tenv)$\;\label{line:simple_post_full}
$\tnotwinningfullskills = \textrm{findNotWinningSkills}(\tfullskills, Z)$\;\label{line:simple_post_notwinning}

\tcp{Find allowable changes}
$\tallowablechanges = \textrm{findAllowableChanges}(\tnotwinningfullskills,\allowbreak \tchangeconstraints,\allowbreak \tnotallowedrepair,\allowbreak \tsysnothard,\allowbreak \tenv, Z)$\;\label{line:simple_post_allowablechanges}
\eIf{$\tallowablechanges \neq \emptyset$}{
$\tselectedchange = \textrm{randomlySelectOneChange}(\tallowablechanges)$\;\label{line:simple_post_selectedchange}
$\tselectedchange = \textrm{removeDependenceOnUserPropositions}(\allowbreak\tselectedchange,\allowbreak \inpuser)$\;\label{line:simple_post_alluservariables}
}
{\Return $\tenv$}

\tcp{Modify $\tenv$}
$\tenvadd = \textrm{findEnvTransitionsToAdd}(\tselectedchange)$\;\label{line:simple_post_tenvadd}
$\tenvold = \textrm{findEnvTransitionsToRemove}(\tselectedchange)$\;\label{line:simple_post_tenvold}
$\tenvnew = (\tenv \setminus \tenvold) \cup \tenvadd$\;\label{line:simple_post_tenvnew}

\Return $\tenvnew$
\caption{\textbf{modify\_postconditions}}
\label{alg:modifypost_simple}
\end{algorithm}

\begin{figure}
	\centering
	\includegraphics[width=\columnwidth]{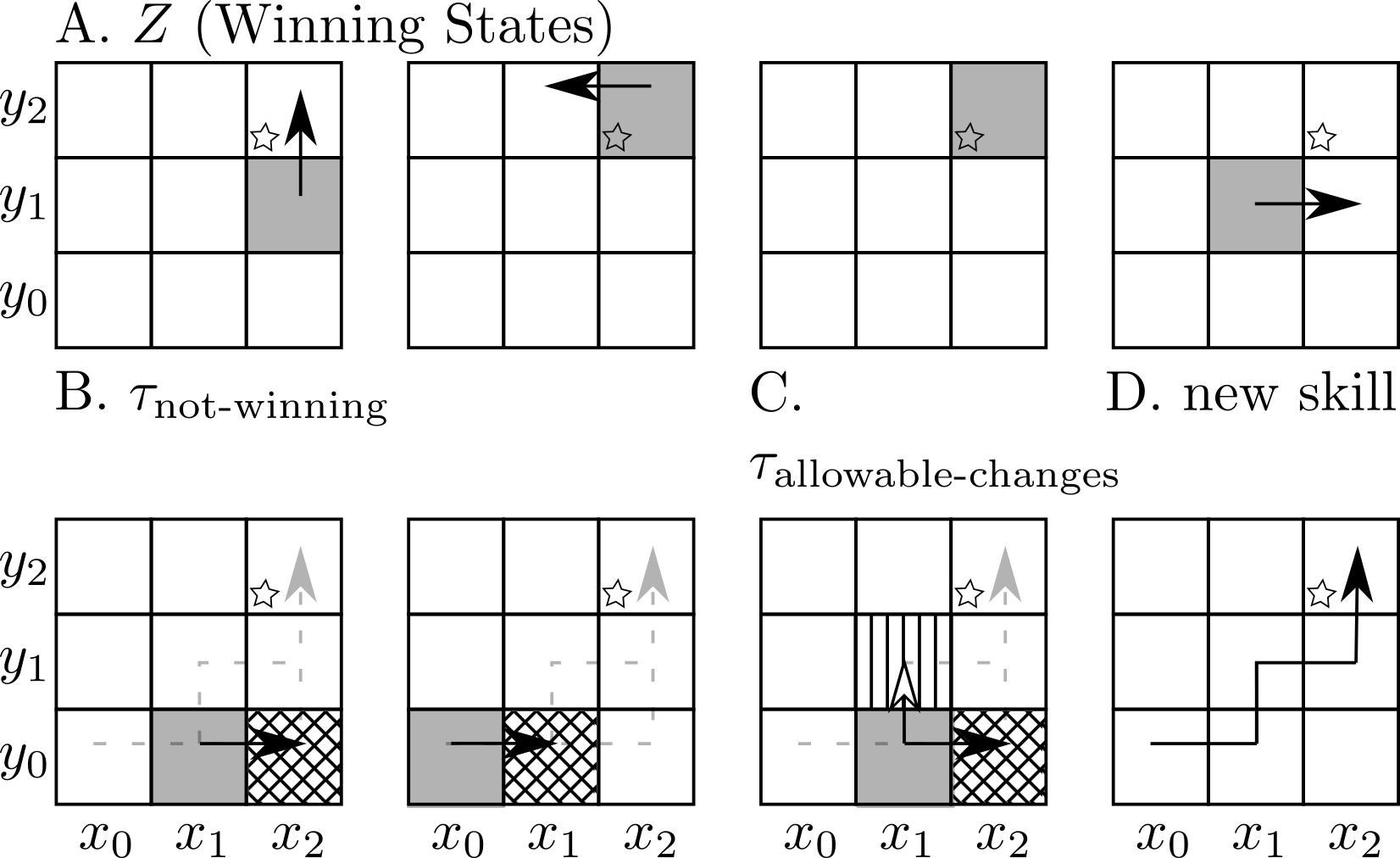}
	\caption{Modifying postconditions in Example~\ref{exm:9s} assuming the change in Fig.~\ref{fig:nine_squares_modify_pre}D has been selected.
	(A) The new winning states $Z$ when the system is trying to satisfy the liveness $\square \lozenge (x_2 \wedge y_2)$.
	The input states are the shaded squares and the arrows are the skill being applied.
	(B) Transitions that are currently not winning. 
	The preconditions are shown by the shaded squares, the current skill with the thick arrow, and the postcondition with the double hashes.
	(C) Allowable changes to the postconditions are shown with vertical hashes.
	The arrow with the white head shows the possible new transition.
	(D) The new skill after modifying the postconditions.
}
	\label{fig:nine_squares_modify_post}
\end{figure}

When modifying postconditions, we automatically change skills that do not lead to a winning state to have postconditions that result in states that are winning, subject to $\changecons$ and $\notallowedrepair$.
This process might reduce nondeterminism (though not in all cases), and in contrast to \cite{pacheck2020finding}, we do not restrict ourselves to only removing postconditions that are not winning, but also allow adding new ones. 
We show the process of modifying the postconditions in Alg.~\ref{alg:modifypost_simple} and in more detail in Alg.~\ref{alg:modifypost} in the Appendix.
In this section, all line references are to Alg.~\ref{alg:modifypost_simple}.
Fig.~\ref{fig:nine_squares_modify_post} shows the process of modifying the postconditions in Example~\ref{exm:9s} after modifying the preconditions as described in Sec.~\ref{sec:modify_pre} and Fig.~\ref{fig:nine_squares_modify_pre}.

In Example~\ref{exm:9s}, after modifying the preconditions, we attempt to synthesize a winning strategy, but the repair process is again triggered when trying to reach $\{x_2, y_2\}$.
Since we modified the preconditions, there are more states from which we can reach $\{x_2, y_2\}$ and $Z=\{\{x_2,\allowbreak y_1,\allowbreak \var_\texttt{L2R}\},\allowbreak \{x_2,\allowbreak y_2,\allowbreak \var_\texttt{R2L}\},\allowbreak \{x_2,\allowbreak y_2\}, \allowbreak\{x_1, \allowbreak y_1, \allowbreak \var_\texttt{L2R}\}\}$ as shown in Fig.~\ref{fig:nine_squares_modify_post}A.
From these states, we are either already in $\{x_2, y_2\}$ or are guaranteed to reach it at some future point.

We first find all of the transitions whose preconditions are reachable
(Line~\ref{line:simple_post_full}) and then identify those skills that neither  start in nor result in $Z$ ($\tnotwinningfullskills$, Line~\ref{line:simple_post_notwinning}).
Fig.~\ref{fig:nine_squares_modify_post}B depicts transitions in $\tnotwinningfullskills$ that could result in task failure due to nondeterminism of the skill.
The components of $\texttt{L2R}$ that are not winning are: $\{x_1,\allowbreak y_0,\allowbreak \var_\texttt{L2R},\allowbreak x_2',\allowbreak y_0',\allowbreak \var_\texttt{L2R}'\}$ and $\{x_0,\allowbreak y_0,\allowbreak \var_\texttt{L2R},\allowbreak x_1',\allowbreak y_0',\allowbreak \var_\texttt{L2R}'\}$.
For example, $\{x_0,\allowbreak y_0,\allowbreak \var_\texttt{L2R},\allowbreak x_1',\allowbreak y_0',\allowbreak \var_\texttt{L2R}'\}$ results in $\{x_1, y_0\}$ at the next step, and must continue executing $\texttt{L2R}$, which may cause the system to reach $\{x_2, y_0, \var_\texttt{L2R}\}$ (which violates the safety constraint and is thus not in the winning set $Z$).

Once we have all the transitions that are not winning, we find how we can change them so they are winning.
We find all allowable changes ($\tallowablechanges$, Line~\ref{line:simple_post_allowablechanges}), making sure they satisfy $\tchangecons$, $\tnotallowedrepair$, $\tsysnothard$, do something, and are winning.
We only allow modifications that either reduce nondeterminism or result in postconditions that are not already states in the skill.
From $\tallowablechanges$, we randomly select one change ($\tselectedchange$, Line~\ref{line:simple_post_selectedchange}).

Fig.~\ref{fig:nine_squares_modify_post}C shows an allowable change for $\texttt{L2R}$; we can change the postcondition  $\{x_2,y_0\}$ to $\{x_1, y_1\}$.
Fig.~\ref{fig:nine_squares_modify_post}D shows the resulting skill after making the modification in Fig.~\ref{fig:nine_squares_modify_post}C. 

Similar to Alg.~\ref{alg:modifyPre_simple}, if there are user controlled propositions, (i.e. $\inpuser \neq \emptyset$), we allow all possible combinations of user controlled variables in the selected modification (Line~\ref{line:simple_post_alluservariables}).

After selecting a change to make to the postconditions, we modify $\tenv$ by removing the transitions associated with the original postcondition and adding the transitions associated with the modified postcondition  (Lines~\ref{line:simple_post_tenvadd}-\ref{line:simple_post_tenvnew}).
If the modification in Fig.~\ref{fig:nine_squares_modify_post}D is chosen, we remove the transition $\{x_1,\allowbreak y_0,\allowbreak \var_\texttt{L2R}, x_2', \allowbreak y_0'\}$ and add the transition $\{x_1,\allowbreak y_0,\allowbreak \var_\texttt{L2R}, x_1', \allowbreak y_1'\}$ to $\tenv$.
Note that in this example, due to the precondition modification, $\{x_1,\allowbreak y_0,\allowbreak \var_\texttt{L2R}, x_1', \allowbreak y_1'\}$ is already part of $\tenv$, so adding it has no effect and the meaningful change is removing a transition.

\subsection{Extracting a Strategy and Modifications to Skills}

The symbolic repair may provide several skill modifications; it may be the case that not all are required for task completion. 
To physically implement only required new skills, when the repair is successful, i.e. the task is realizable given the original and new skills, we extract the required symbolic changes to the skills by finding a deterministic strategy and comparing $\tenv$ and $\tsysnothard$ of the deterministic strategy to the original $\tenv$ and $\tsysnothard$. 
This skill is then checked for physical feasibility. 

For Example~\ref{exm:9s}, when selecting the change in Fig.~\ref{fig:nine_squares_modify_post}D, we find we need a new skill based on $\texttt{L2R}$ with intermediate states as shown in Eq.~\eqref{eq:intermediate_states_l2r_new}.
\begin{equation}\label{eq:intermediate_states_l2r_new}
    \begin{aligned}
        \dom^\textrm{pre-init}_\texttt{L2R} &= \{\{x_0,y_0\}\}
        &\dom_{\texttt{L2R},\{x_0,y_0\}}^\textrm{post} &= \{\{x_1,y_0\}\}\\
        \dom_{\texttt{L2R},\{x_1,y_0\}}^\textrm{post} &= \{\{x_1,y_1\}\}
        &\dom_{\texttt{L2R},\{x_1,y_1\}}^\textrm{post} &= \{\{x_2,y_1\}\}\\
        \dom_{\texttt{L2R},\{x_2,y_1\}}^\textrm{post} &= \{\{x_2,y_2\}\}
        &\dom^\textrm{final-post}_\texttt{L2R} &= \{\{x_2,y_2\}\}
    \end{aligned}
\end{equation}

\section{Physical Implementation of Skills}\label{sec:physicalimplementation}

Once we have symbolic suggestions for skill modifications, we determine whether the skills are physically implementable. 
In this work we treat skills as trajectories generated by dynamic motion primitives; however, other approaches would also work (e.g. \cite{jaquier2019learning, noseworthy2019task, kulak2020fourier}).

The first step is finding a new skill that has trajectories as close as possible to the original skill while satisfying the new constraints. 
If we find such a skill, we check whether the robot can physically follow the resulting trajectories.
If, at either step, we find that the new skill is not possible, we determine which parts of the skill are not feasible and add constraints to the symbolic repair process as described in Sec.~\ref{sec:feedback}.

There are multiple methods to find modifications to skills given symbolic constraints (e.g. \cite{innes2020elaborating, rana2018towards}).
For this work, we use the method proposed in \cite{innes2020elaborating}; it transforms an \gls{ltl} constraint into a loss function.
Together with the initial skill, this constraint allows us to generate a new \gls{dmp}.

To use the method in \cite{innes2020elaborating}, we write each skill suggestion as an \gls{ltl} formula in the form of conjunction of implications as shown in Eq.~\eqref{eq:physical_ltl}. 
The formula requires skills to always go from a precondition to either the same precondition or one of the postconditions and always stay in the unique states of the skill.
In the suggestion for Example~\ref{exm:9s} $\dom_\texttt{L2R}^\textrm{unique} = \{\{x_0, y_0\},\allowbreak \{x_1, y_0\}, \allowbreak \{x_1, y_1\}, \allowbreak \{x_2, y_1\}, \allowbreak \{x_2, y_2\}\}$.
Eq.~\eqref{eq:nine_squares_suggestion} in the Appendix encodes the new constraints on the skill. 
\begin{equation}\label{eq:physical_ltl}
    \begin{split}
        &\bigwedge_{\statevar_\textrm{pre} \in \dom_\skill^\textrm{unique} \setminus \dom_\skill^\textrm{final-post}} \square \left(\varphi_{\statevar_\textrm{pre}} \rightarrow \bigcirc (\varphi_{\statevar_\textrm{pre}} \vee \bigvee_{\statevar_\textrm{post} \in \dom^\textrm{post}_{\skill,\statevar_\textrm{pre}}} \varphi_{\statevar_\textrm{post}}\right)\\
        &\wedge
        \square \left(\bigvee_{\statevar_\textrm{unique} \in \dom_\skill^\textrm{unique}} \varphi_{\statevar_\textrm{unique}} \right)
    \end{split}
\end{equation}

To modify the skill, we take the original skill and the \gls{ltl} constraint generated from the symbolic repair and train a neural network that generates the weights for a \gls{dmp} that outputs the trajectory of the modified skill.
The method proposed by \cite{innes2020elaborating} allows the user to tune the trade off between remaining close to the original demonstrations versus obeying the \gls{ltl} constraints.
In this work, we weighted satisfying the \gls{ltl} constraints 50 times more than remaining close to the original demonstrations.
For each skill suggestion, we perform 100 epochs of training to create the neural network that generates the weights for the \gls{dmp}.
The second column of Fig.~\ref{fig:nine_squares_a_physical} shows the original trajectories for the skill $\texttt{L2R}$ in Example~\ref{exm:9s}.
The third and fourth columns of Fig.~\ref{fig:nine_squares_a_physical}C show the modified trajectories generated by the \gls{dmp} subject to the \gls{ltl} constraint generated from the suggestion in Eq.~\eqref{eq:intermediate_states_l2r_new} and based off the original skill.
The third column of Fig.~\ref{fig:nine_squares_a_physical}C shows the trajectories that satisfy the constraint and the fourth column shows the trajectories that do not.

\section{Feedback between Symbolic and Physical Repair}\label{sec:feedback}

Crucial and unique to our approach is the tight integration between the symbolic and physical repair. 
As described in the previous sections, the symbolic repair provides constraints to the physical repair; in this section we describe the constraints the physical repair provides the symbolic. 

After attempting to physically implement the modified skill (Sec.~\ref{sec:physicalimplementation}), we check whether it satisfies the new constraints.
We do this in two steps: first we gain confidence that the new skill satisfies the constraints, by sampling trajectories and ensuring the percentage of trajectories that satisfy the constraints (Eq.~\eqref{eq:physical_ltl}) is above a threshold, here 90\%. 
Second, we verify that there exists a controller for the robot that can implement the trajectory.
For example, for the Baxter  (Sec.~\ref{sec:demo_baxter}), we check that the sampled end-effector trajectories have valid inverse kinematics solutions.
If the new trajectory satisfies the constraints and a controller exists, the new skill is used to repair the task.
For the (simulated) Nine Square example (Example~\ref{exm:9s}), we assume a holonomic point robot that can reliably follow any trajectory, therefore we only consider the trajectory validity.

If a physically feasible trajectory was not found, we determine which transitions are physically unlikely and use those transitions to restrict future symbolic suggestions, by adding them to $\notallowedrepair$.
Specifically, for each intermediate transition of the skill, we create a monitor that checks the formula in Eq.~\eqref{eq:monitor} -- whether for each precondition, the trajectory either stays in the precondition or moves to an appropriate postcondition  (based on Eq.~\eqref{eq:physical_ltl}). 
\begin{equation}\label{eq:monitor}
    \varphi^\textrm{monitor} = \square \left(\varphi_{\statevar_\textrm{pre}} \rightarrow \bigcirc\left(\varphi_{\statevar_\textrm{pre}} \vee \bigvee_{\statevar_\textrm{post} \in \dom^\textrm{post}_{\skill,\statevar_\textrm{pre}}} \varphi_{\statevar_\textrm{post}} \right) \right)
\end{equation}
For each transition that did not robustly succeed, we add $\square \lnot(\varphi_{\statevar_\textrm{pre}} \wedge \var_\skill \wedge \bigcirc \bigvee_{\statevar_\textrm{post} \in \dom_{\skill, \statevar_\textrm{pre}}^\textrm{post}} \varphi_{\statevar_\textrm{post}})$ to $\notallowedrepair$.

We show the integrated symbolic-physical repair process in Fig.~\ref{fig:nine_squares_a_physical}.
The first two rows show physically impossible symbolic suggestions, the attempted physical implementation, and the additions to $\tnotallowedrepair$.
After forbidding those transitions, we get a physically possible symbolic suggestion as shown in the third row of Fig.~\ref{fig:nine_squares_a_physical}.

\section{Simulated Demonstrations}\label{sec:demos}

\begin{figure}
    \centering
    \includegraphics[width=\columnwidth]{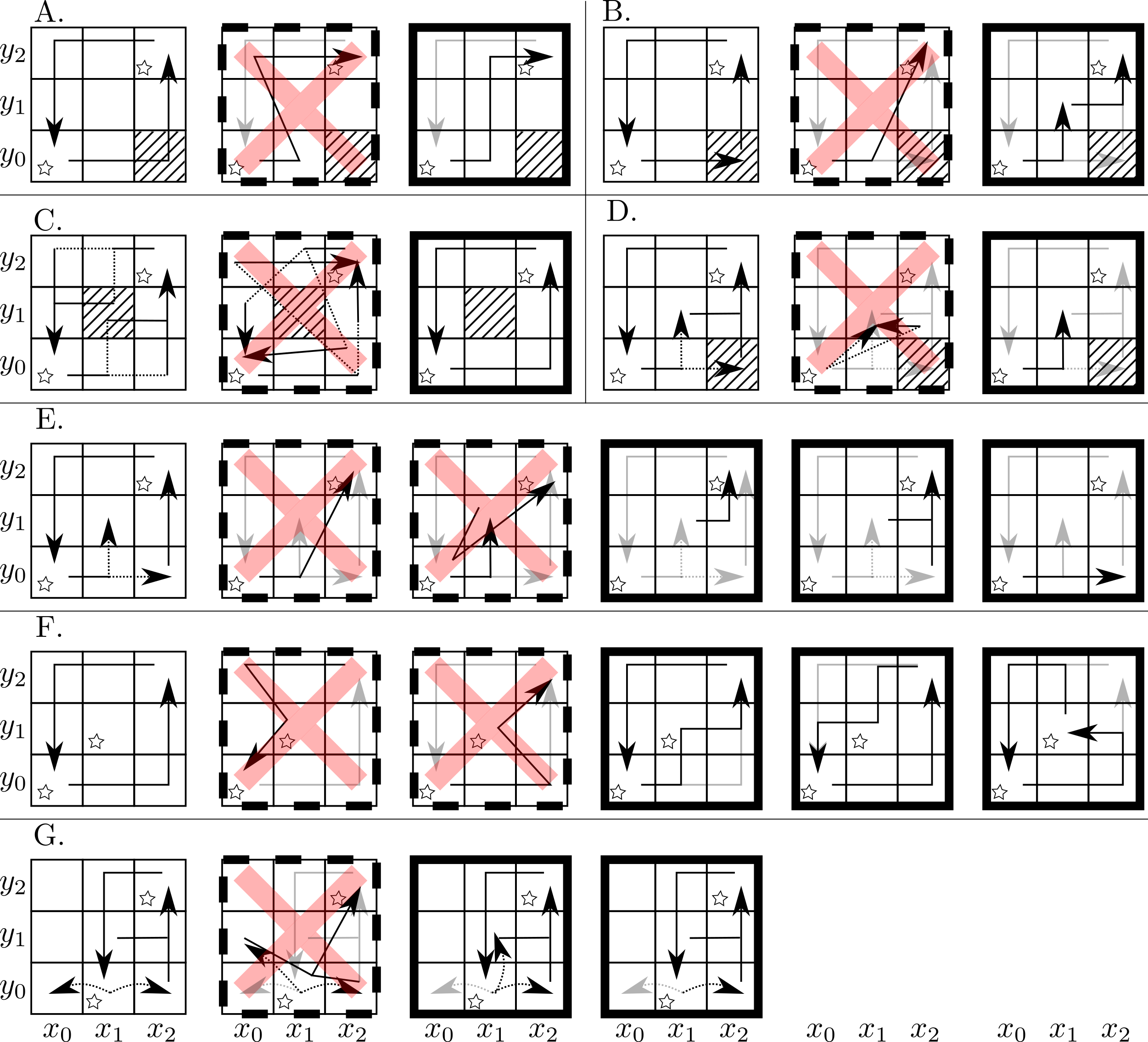}
    \caption{Different skills and repairs. 
    Each subfigure contains the initial skills and task, physically infeasible repairs (dashed squares with red ``x'') and possible repairs (bold squares). 
    Skills with nondeterministic outcomes are shown with dotted lines.
    Modified skills are bold in the suggestions.
    Original skills, if not modified, are shown with a light-colored arrow.
    The task is to avoid the hashed areas and always eventually reach the areas with stars. 
    }
    \label{fig:sym_demos}
\end{figure}

We present several demonstrations showcasing the different capabilities of the repair process, including multiple skills, modifying intermediate states, preconditions, postconditions, and achieving livenesses at both the postcondition and during the skill.
The code for the algorithms is available at: \url{https://github.com/apacheck/synthesis_based_repair}.
Fig.~\ref{fig:sym_demos} illustrates the examples and shows the skills and tasks, together with symbolic suggestions that were found to be physically impossible (squares with a dashed outline and red ``x''), and physically implementable skills that resulted from the repair (squares with a bold outline). 
Each bold square represents one set of suggestions from the repair process, which may include changes to multiple skills. 
Here, each trajectory is modeled as a \gls{dmp} with 50 discrete points.

\begin{figure}
    \centering
    \includegraphics[width=\columnwidth]{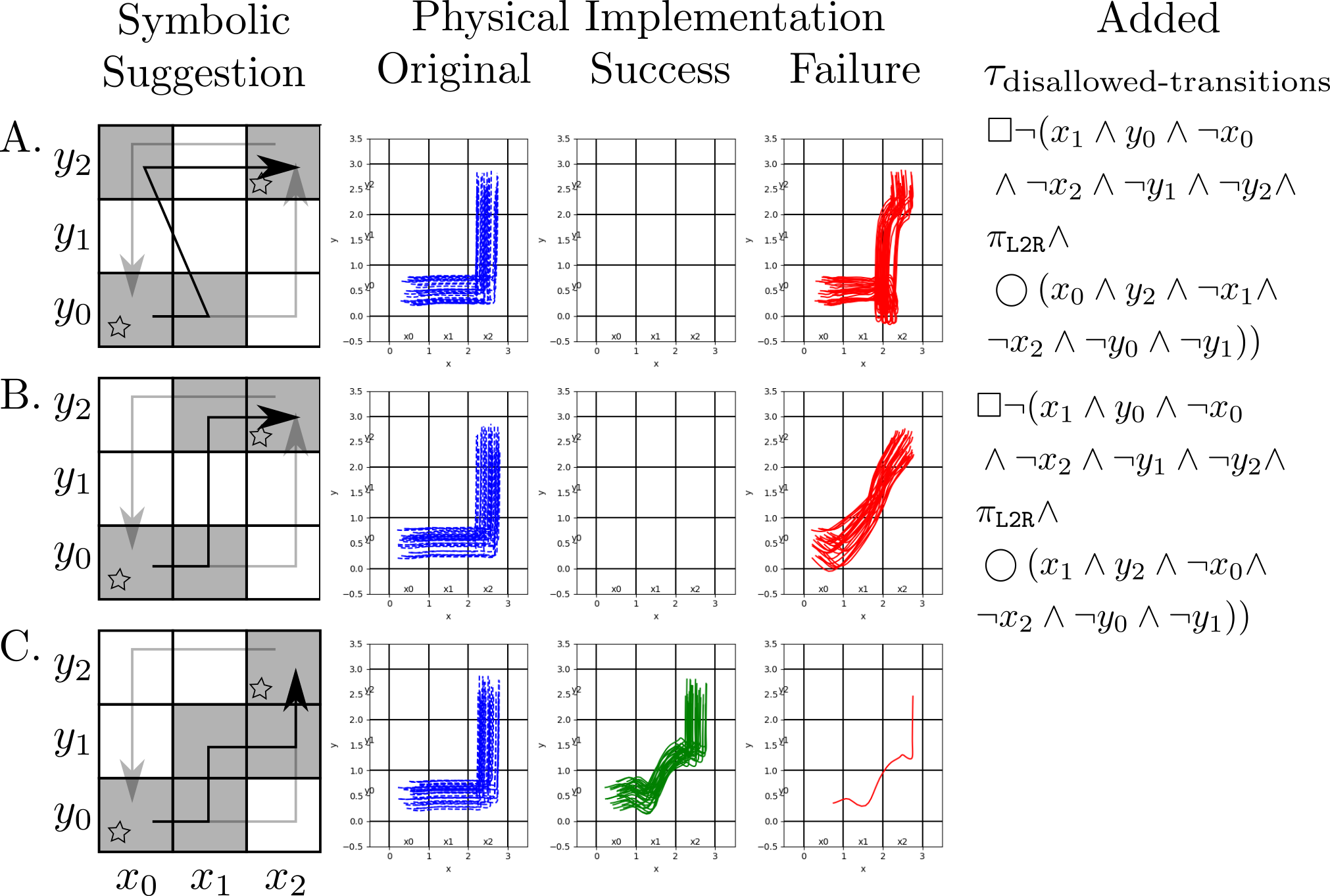}
    \caption{Steps in the symbolic-physical repair process for Example~\ref{exm:9s}. 
    The first column depicts the symbolic suggestions.
    The gray shaded squares are the symbolic states the skill should pass through following the black arrow.
    The second column shows sampled trajectories from  the original skill.
    The third column shows sampled trajectories that successfully implement the skill modification and the fourth shows the trajectories that failed.
    The final column shows the constraint added to $\tnotallowedrepair$.}
    \label{fig:nine_squares_a_physical}
\end{figure}

\noindent\textit{A) Modify Intermediate States:}
For Example~\ref{exm:9s} ( $\syslive=\square \lozenge (x_0 \wedge y_0) \wedge \square \lozenge (x_2 \wedge y_2)$ and $\syssafetyuser = \square \lnot (x_2 \wedge y_0) \wedge \square \lnot \bigcirc (x_2 \wedge y_0)$), one repair is depicted in  Fig.~\ref{fig:nine_squares_a_physical}, and another in Fig.~\ref{fig:sym_demos}A.

\noindent\textit{B) Modify Two Skills:}
Fig.~\ref{fig:sym_demos}B shows the same task as the previous one, but with three skills instead of two.
The repair process suggests physically implementable modifications to two skills. 
These suggestions modify the final postconditions for one skill and the initial preconditions for the other. 

\noindent\textit{C) Modify Two Skills Intermediate States:}
Fig.~\ref{fig:sym_demos}C shows an example where the hard constraint is to avoid $\{(x_1, y_1)\}$ and the liveness is the same as in Example~\ref{exm:9s}.
The repair process finds suggestions that modify the intermediate states, eliminating nondeterminism.

\noindent\textit{D) Modify Postconditions:}
In Fig.~\ref{fig:sym_demos}D, the repair process suggests modifying the postconditions of the skill with nondeterministic outcomes.
Note that the skill with a precondition that includes two states does not need to be modified.

\noindent\textit{E) Multiple Different Repairs:} Our repair process is able to find multiple physically possible suggestions, if they exist, as shown in Fig.~\ref{fig:sym_demos}E.
The task has the same liveness as Example~\ref{exm:9s} but there are no user-added safety constraints.
The task can be repaired by modifying either the initial preconditions or final postconditions of the original skills.

\noindent\textit{F) Intermediate Liveness:}
Fig.~\ref{fig:sym_demos}F depicts a task with $\syslive = \square \lozenge (x_0 \wedge y_0) \wedge \square \lozenge (x_1 \wedge y_1)$ where the repair suggestions involve modifying the intermediate states of a skill such that the task is satisfied during the execution of a skill, in contrast to prior work where task completion was only determined after skill execution.
As shown in the fourth and fifth columns of Fig.~\ref{fig:sym_demos}F, two possible suggested modifications are to modify one of the intermediate states of either skill such that it passes through $\{x_1,y_1\}$.
Modifying the final postcondition of one of the skills to end in $\{x_1,y_1\}$ and the precondition of the other skill to start in $\{x_1,y_1\}$ is another possible suggestion.

\noindent\textit{G) Nondeterministic Final Postcondition:}
We show an example involving a skill with nondeterministic final postconditions in Fig.~\ref{fig:sym_demos}G that does not necessarily reduce the nondeterminism, but can change it instead.
The postcondition that ends in $\{x_0, y_0\}$ can either be changed to $\{x_1, y_1\}$ or $\{x_2, y_0\}$.
The modification to $\{x_1, y_1\}$ does not remove nondeterminism, but simply changes it, while the change to $\{x_2, y_0\}$ reduces nondeterminism.

\noindent\textit{H) User-Controlled Inputs:}
Our repair process can also handle specifications with user controlled propositions.
Given a task $\syslive = \square \lozenge (\var_\textrm{react} \rightarrow x_2 \wedge y_2) \wedge \square \lozenge (\lnot \var_\textrm{react} \rightarrow x_0 \wedge y_0)$ and the skills in Fig.~\ref{fig:sym_demos}A, we perform repair.
The skill suggestions are the same as in Example~\ref{exm:9s} and Fig.~\ref{fig:sym_demos}A.

\section{Physical Demonstrations}
We demonstrate the generality of our approach by presenting two different physical robots (Baxter and Jackal), their abstractions, and tasks.
We discuss what the physical implementation of modification entails for these systems and what the process is when moving to a new physical system. 
\subsection{Baxter moving a block}\label{sec:demo_baxter}

\begin{figure}
    \centering
    \includegraphics[width=\columnwidth]{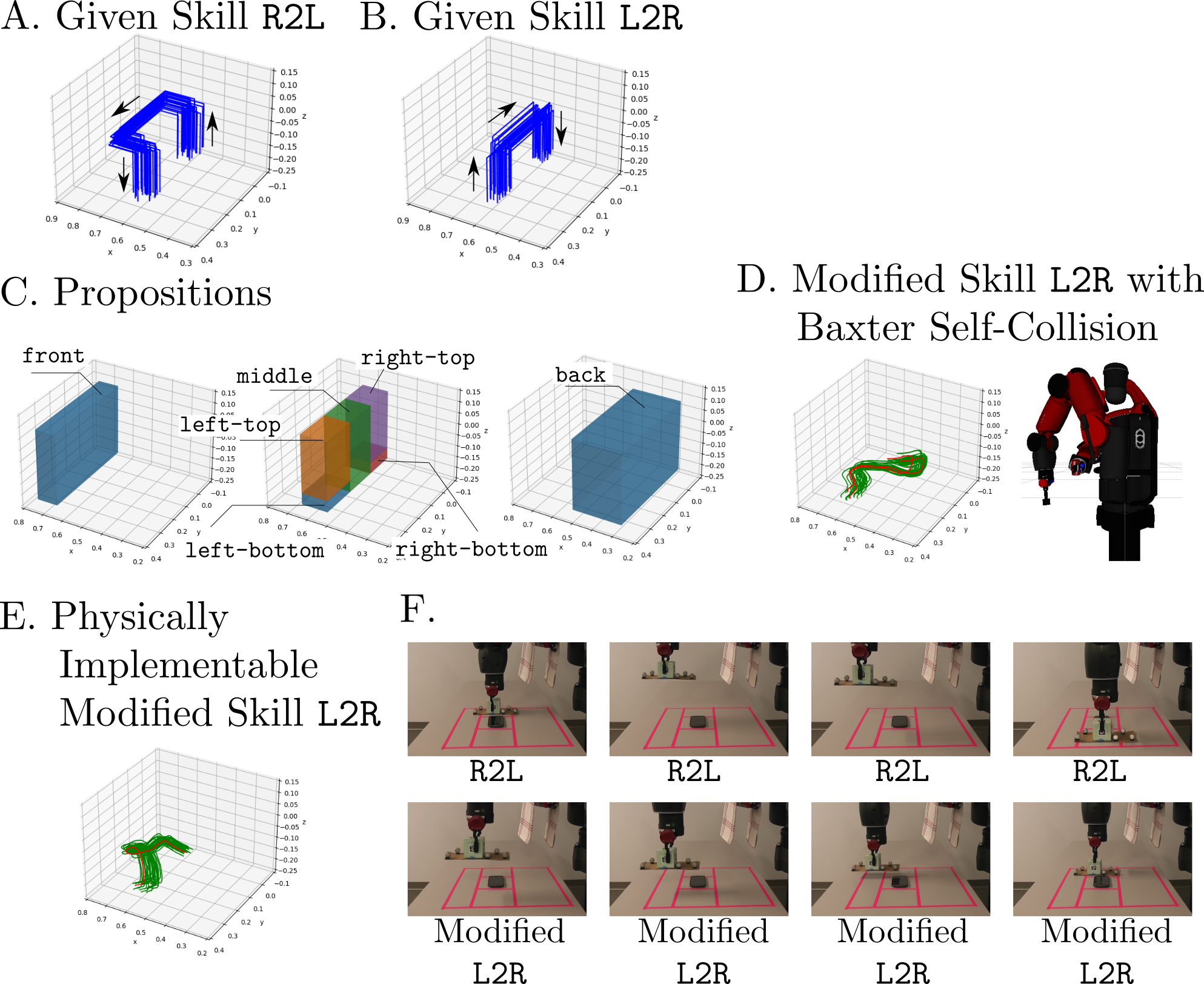}
    \caption{Baxter moving a block.
    The task is $\syslive = \square \lozenge \texttt{right-bottom} \wedge \square \lozenge \texttt{left-bottom}$ and $\syssafetyuser = \square \lnot \texttt{middle} \wedge \square \lnot \bigcirc \texttt{middle}$.
    A) Initial skill \texttt{R2L}.
    B) Initial skill \texttt{L2R}.
    C) Propositions grounding.
    D) Physical implementation of the trajectory for a suggestion modifying the skill \texttt{L2R} to pass through \texttt{back} instead of \texttt{middle}.
    A valid trajectory for the block exists, but the robot arm cannot follow that trajectory due to self-collisions.
    E) Successful physical implementation of the trajectory for a suggestion modifying the skill \texttt{L2R} to pass through \texttt{front} instead of \texttt{middle}.
    F) Images from the Baxter successfully performing the task with the modified skill.}
    \label{fig:baxter_example}
\end{figure}

Here, the Baxter robot is moving a block on the table (Fig.~\ref{fig:baxter_example}).
The task is to repeatedly move the block from one side of the table to the other while avoiding the phone in the middle of the table.

\noindent\textbf{Abstraction:} 
We abstract the position of the green block using $\inpsyms = \{\var_\texttt{left-bottom},\allowbreak \var_\texttt{left-top},\allowbreak \var_\texttt{right-bottom},\allowbreak \var_\texttt{right-top},\allowbreak \var_\texttt{front},\allowbreak \var_\texttt{back},\allowbreak \var_\texttt{middle}\}$.
Each proposition is grounded to a rectangular region of space such that $\forall \pi_i, \pi_j, \; G(\pi_i)\cap G(\pi_j)=\emptyset$ (Fig.~\ref{fig:baxter_example}C).
We determine the position of the block using an OptiTrack motion capture system. 
There are no user controlled propositions (i.e. $\inpuser = \emptyset$).

\noindent \textbf{Skills:} The Baxter has two skills, $\texttt{L2R}$ and $\texttt{R2L}$ (Fig.~\ref{fig:baxter_example}A-B).
The skill $\texttt{L2R}$ goes from $\var_\texttt{left-bottom}$ to $\var_\texttt{left-top}$ to $\var_\texttt{middle}$ to $\var_\texttt{right-top}$ to $\var_\texttt{right-bottom}$ and skill $\texttt{R2L}$ goes from $\var_\texttt{right-bottom}$ to $\var_\texttt{right-top}$ to $\var_\texttt{front}$ to $\var_\texttt{left-top}$ to $\var_\texttt{left-bottom}$.
We created these skills by defining training trajectories as shown in Fig.~\ref{fig:baxter_example}A and B. 
For $\texttt{L2R}$, we sample 100 waypoints from locations grounded to $\var_\texttt{left-bottom}$, $\var_\texttt{left-top}$, $\var_\texttt{right-top}$, and $\var_\texttt{right-bottom}$ to define the trajectories.
The point grounded to $\var_\texttt{left-top}$ is constrained to have the same $x$ and $y$ coordinates as the point grounded to $\var_\texttt{left-bottom}$.
The point grounded to $\var_\texttt{right-top}$ is constrained to have the same $y$ and $z$ coordinates as the point grounded to $\var_\texttt{left-top}$.
The point grounded to $\var_\texttt{right-bottom}$ is constrained to have the same $x$ and $y$ coordinates as the point grounded to $\var_\texttt{right-top}$.
Given these waypoints that define the trajectories, we train the weights for a \gls{dmp} to reproduce the waypoints given a start and end point.
When the robot executes a skill it finds the joint angles needed to reach each point in the trajectory (using Trak-IK inverse kinematics solver \cite{beeson2015trac}) and then moves between the joint angles.

\noindent \textbf{Physical Feasibility:} To automatically validate physical feasibility, for the Baxter we consider two aspects: existence of an inverse kinematic solution, and no self collisions. Given a trajectory (provided by the \gls{dmp} that was modified to satisfy the symbolic repair) we use the Trak-IK inverse kinematics solver \cite{beeson2015trac} to find the joint angles to achieve the points on the trajectory and then check that the proposed joint angles do not cause self-collision with the body or other arm.

\noindent \textbf{Task:} The task is to move the block between $\var_\texttt{right-bottom}$ and $\var_\texttt{left-bottom}$ while avoiding $\var_\texttt{middle}$ (i.e. $\syslive = \square \lozenge \var_\texttt{right-bottom} \wedge \square \lozenge \var_\texttt{left-bottom}$ and $\syssafetyuser = \square \lnot \var_\texttt{middle} \wedge \square \lnot \bigcirc \var_\texttt{middle}$).

\noindent \textbf{Repair:} The task is not possible with these two skills.
The symbolic repair first suggests a modification to $\texttt{L2R}$ such that instead of passing through $\var_\texttt{middle}$, it passes through $\var_\texttt{back}$ and immediately goes to $\var_\texttt{right-bottom}$.
The robot is able to find a trajectory for the block that satisfies these new constraints (Fig.~\ref{fig:baxter_example}D).

For this skill, our process determines that for the proposition $\var_\texttt{back}$ to be $\true$, the arm needs to move too close to the Baxter's torso and thus there are self-collisions.
We therefore automatically add $\square \lnot (\var_\texttt{left-top} \wedge \var_\texttt{L2R} \wedge \bigcirc \var_\texttt{back})$ and $\square \lnot (\var_\texttt{back} \wedge \var_\texttt{L2R} \wedge \bigcirc \var_\texttt{right-bottom})$ to $\notallowedrepair$.

Continuing the repair, the new suggestion is to modify $\texttt{L2R}$ such that it passes through $\texttt{front}$ instead of $\texttt{middle}$ and then goes directly to $\var_\texttt{right-bottom}$.
Our automated physical feasibility checking determines that the skill can be executed, as shown in Fig.~\ref{fig:baxter_example}E.
Fig.~\ref{fig:baxter_example}F shows the Baxter using this modified skill to accomplish the task.

\subsection{Jackal patrolling and reacting to input from a person}\label{sec:demo_jackal}

We show a demonstration with a Jackal robot performing a patrolling task.
Here we highlight two differences from the previous examples:  the robot is reacting to input from a person, and the abstraction does not create a partition of the continuous state space, i.e. the grounding of some proportions overlap.

\begin{figure}
    \centering
    \includegraphics[width=\columnwidth]{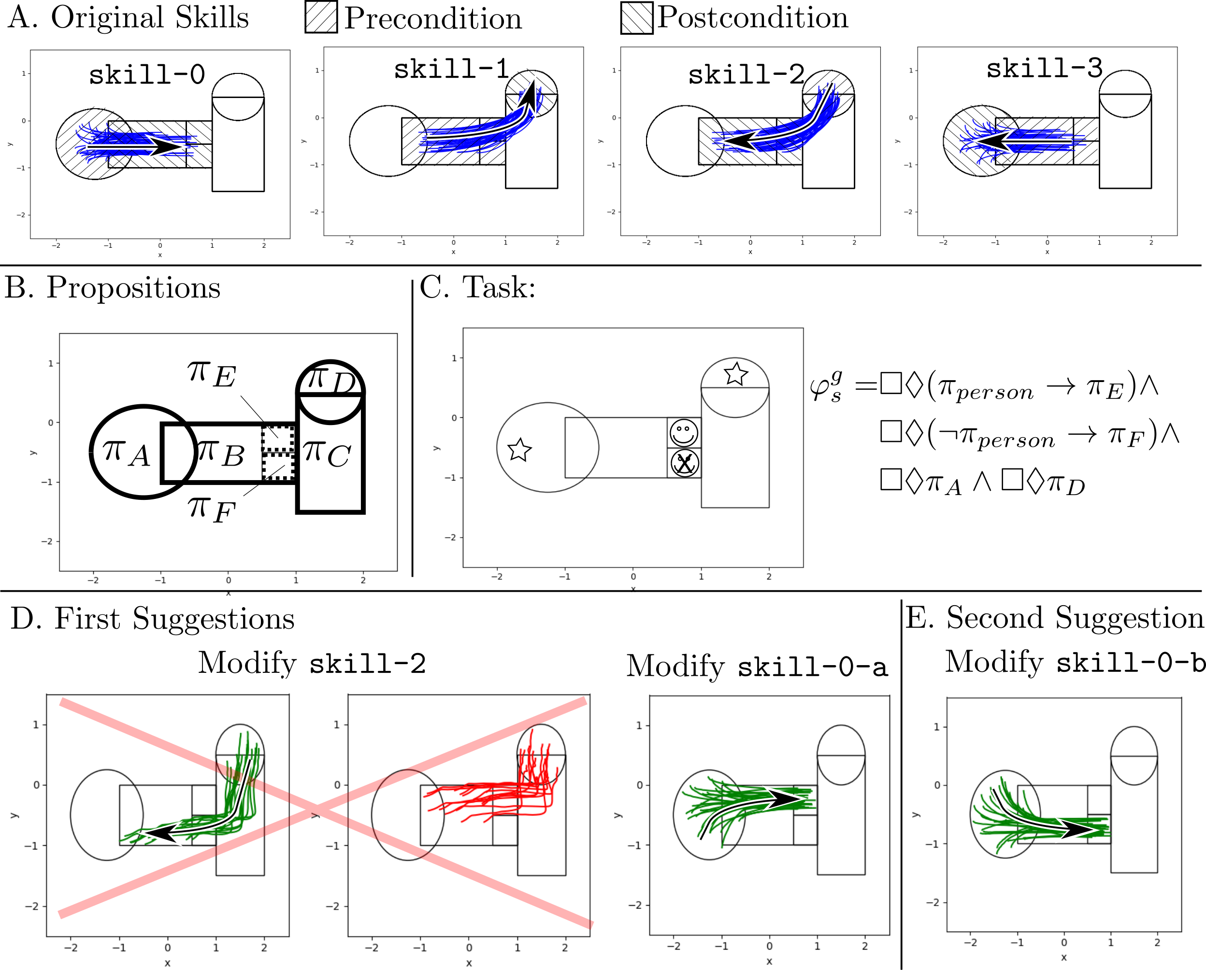}
    \caption{Demonstration with a Jackal robot.
    A) Initial skills.
    B) Propositions.
    C) The task is $\syslive = \square \lozenge (\var_{person} \rightarrow \var_{E}) \wedge \square \lozenge (\lnot \var_{person} \rightarrow \var_{F}) \wedge \square \lozenge \var_{A} \wedge \square \lozenge \var_{D}$
    D) Suggestions and physical implementation.
    The first symbolic suggestions contains modifications to $\texttt{skill-2}$ and $\texttt{skill-0}$, denoted by the black arrows. For $\texttt{skill-2}$ we did not find a suitable physical implementation, while for $\texttt{skill-0}$ we did.
    E) The second suggestion is to create another modification for $\texttt{skill-0}$ such that it only results in $\var_{F}$; this repair was also successful.
    }
    \label{fig:jackal_demo}
\end{figure}

\noindent\textbf{Abstraction:} We abstract the position of the Jackal in the workspace using $\inpsyms = \{\var_A, \var_B, \var_C, \var_D, \var_E, \var_F\}$. 
The propositions are grounded to either rectangles in (X,Y) or circles. 
Furthermore, the propositions do not represent a partition of the workspace as in previous examples but rather show that the approach can be applied to overlapping groundings (Fig.~\ref{fig:jackal_demo}B).
Note that $\var_E$ and $\var_F$ are squares contained in the rectangle $\var_B$.
We tracked the position of the Jackal using a Vicon motion capture system.
In addition, the user controls $\inpuser = \{\var_\textrm{person}\}$ via a computer interface.

\noindent \textbf{Skills:} The robot has four skills.
Each skill is generated by randomly sampling pairs of points from the precondition and postcondition and generating a curved trajectory between them.
The weights of the \gls{dmp} corresponding to the skill are learned from these initial trajectories.
Fig.~\ref{fig:jackal_demo}A shows the preconditions, postconditions, and original trajectories.
For example, $\texttt{skill-0}$ has precondition of $\var_{A}$ and postcondition of $\var_{B}$ (i.e. $\Sigma_{\texttt{skill-0}}^\textrm{pre-init} = \{\{\var_A\}, \allowbreak \{\var_A, \var_B\}\}$, $\Sigma_{\texttt{skill-0}}^\textrm{final-post} = \{\{\var_B\}, \allowbreak \{\var_A, \var_B\}, \allowbreak \{\var_B, \var_E\}, \allowbreak \{\var_B, \var_F\}\}$).
After executing $\texttt{skill-0}$, $\var_B$ will always be $\true$, but $\var_A$, $\var_E$, or $\var_F$ may be $\true$ as well, due to the overlap in the grounding sets.
When a skill is executed, a point is randomly sampled from within the postconditions and passed to the \gls{dmp} along with the current pose of the robot to generate the desired trajectory. % for the robot.

\noindent \textbf{Physical Feasibility:}
To physically implement the skill, we control the Jackal to follow the waypoints of the trajectory using feedback linearization.
Due to the small turning radius of the Jackal, we assume it can follow any sequence of waypoints.

\noindent \textbf{Task:} The task is $\syslive = \square \lozenge (\var_{person} \rightarrow \var_{E}) \wedge \square \lozenge (\lnot \var_{person} \rightarrow \var_{F}) \wedge \square \lozenge \var_{A} \wedge \square \lozenge \var_{D}$; the robot should repeatedly visit $\var_A$ and $\var_D$, when the user input is $\true$ it should visit $\var_E$, and when the user input is $\false$ it should visit $\var_F$ (Fig~\ref{fig:jackal_demo}C).
This task does not contain safety constraints, and the requirements are of the form $\square \lozenge$, therefore, for the task to be achieved, the jackal needs to repeatedly, go through $\var_{E}$ when $\var_{person}$ is $\true$ and through $\var_{F}$ when it is $\false$.

This task is unrealizable with the given skills because the robot cannot guarantee, due to the nondeterminism of the skills,  that it will reach $\var_E$ or $\var_F$ to react to the user input.

\noindent \textbf{Repair:} The initial suggestion contains modifications to two skills, $\texttt{skill-2}$ and $\texttt{skill-0}$ (Fig.~\ref{fig:jackal_demo}D).
The repair process suggests modifying $\texttt{skill-2}$ such that it always passes through $\{\var_B, \var_F\}$ and does not pass through $\{\var_B, \var_E\}$.
With the current parameters of weighing the constraints 50 times more than remaining close to the original trajectory and training for 100 epoch for the skill learning, we were unable to find a \gls{dmp} that can generate trajectories that satisfy these constraints for more than 90\% of the initial and final points.
The second modification of the suggestion is for $\texttt{skill-0}$ to always result in $\{\var_B, \var_E\}$, which is physically possible.
The repair process continues, incorporating the successful modification to $\texttt{skill-0}$, and returns another suggestion to modify $\texttt{skill-0}$, this time to result in $\{\var_B, \var_F\}$ (Fig.~\ref{fig:jackal_demo}E).
This modification is also successful.
With these two modified skills based on $\texttt{skill-0}$, we are able to successfully complete the task and react to input from a person. 

\section{Conclusion}

Within the context of creating complex robot behaviors through skill composition, in this work we address two challenges; taking into account intermediate states a skill may go through when physically executed by a robot, and providing an automated process for repairing infeasible tasks when there are missing skills.

We have shown that our approach generalizes to different robots, abstractions and tasks; but in order to do so, we need automated ways of creating controllers that satisfy LTL constraints. 
In this paper we use~\cite{innes2020elaborating}, in the future we will explore other approaches such as the reinforcement learning methods as proposed in~\cite{li2017reinforcement} or the modification of skills learned from demonstration proposed in \cite{rana2018towards}. 
Furthermore, given a robot, it is possible to automate the feasibility checking, for example using an IK solver, as we have shown with the Baxter. 
For other robots, we will explore the properties needed for such feasibility checking and how to suggest suitable techniques.

Another future direction is incorporating optimality criteria in the repair process. 
In this work we do allow the user to influence the symbolic repair by adding constraints on possible modifications. 
In the future, we will explore how continuous metrics can be specified and incorporated in the repair.

\section*{Acknowledgments}
This work is supported by the ONR PERISCOPE MURI award N00014-17-1-2699.

% \clearpage
% \clearpage

% \bibliographystyle{IEEEtran}
% \bibliography{references.bib}

% \newpage

 
\vspace{11pt}

\vspace{-33pt}
\begin{IEEEbiography}[{\includegraphics[width=1in,height=1.25in,clip,keepaspectratio]{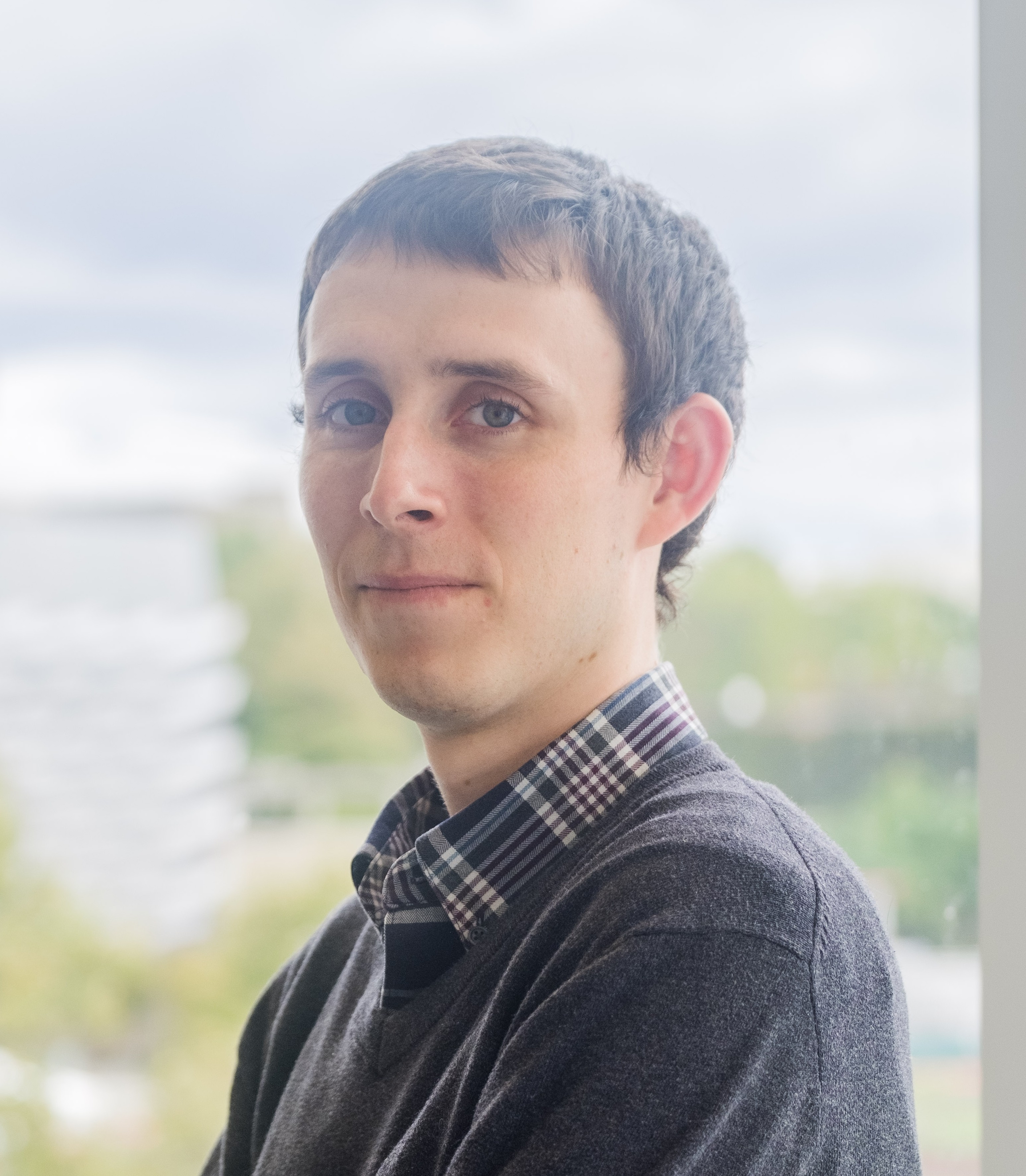}}]{Adam Pacheck}
recieved the Ph.D. degree in mechanical engineering from Cornell University in Ithaca, NY, USA in 2022.

His research focuses on repair of skills to complete high-level tasks.
\end{IEEEbiography}

\vspace{-33pt}
\begin{IEEEbiography}[{\includegraphics[width=1in,height=1.25in,clip,keepaspectratio]{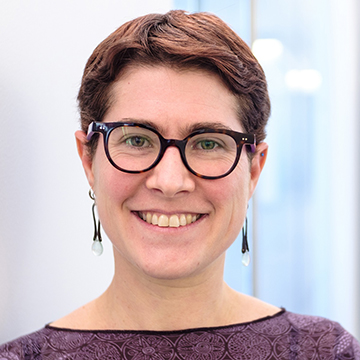}}]{Hadas Kress-Gazit} is the Geoffrey S.M. Hedrick Sr. Professor at the Sibley School of Mechanical and Aerospace Engineering at Cornell University. She received her Ph.D. in Electrical and Systems Engineering from the University of Pennsylvania in 2008 and has been at Cornell since 2009. Her research focuses on formal methods for robotics and automation and more specifically on synthesis for robotics – automatically creating verifiable robot controllers for complex high-level tasks. Her group explores different types of robotic systems including modular robots, soft robots and swarms and synthesizes ideas from robotics, formal methods, control, hybrid systems, and computational linguistics. She is an IEEE fellow and has received multiple awards for her research, teaching and advocacy for groups traditionally underrepresented in STEM.
\end{IEEEbiography}
\vfill

% \newpage
% \newpage

% \clearpage
% \newpage
\appendix
% \clearpage
The appendix provides formulas demonstrating the encoding of the skills and modifications in \gls{ltl} formulas.
Additionally, we provide more detailed versions of Alg.~\ref{alg:modifyPre_simple} and Alg.~\ref{alg:modifypost_simple}.
Postconditions of \texttt{L2R} in $\envsafetynothard$ for Example~\ref{exm:9s} (Eq.~\ref{eq:env_trans}): 
    \begin{equation}\label{eq:tenv_lefttoright}
        \begin{split}
            &\square ((\var_\texttt{L2R} \wedge x_0  \wedge y_0 \wedge \lnot x_1 \wedge \lnot x_2 \wedge \lnot y_1 \wedge \lnot y_2) \rightarrow \\
            &\bigcirc ( x_1  \wedge y_0 \wedge \lnot x_0 \wedge \lnot x_2 \wedge \lnot y_1 \wedge \lnot y_2)) \wedge\\
            &\square((\var_\texttt{L2R} \wedge x_1  \wedge y_0 \wedge \lnot x_0 \wedge \lnot x_2 \wedge \lnot y_1 \wedge \lnot y_2) \rightarrow \\
            &\bigcirc (x_2 \wedge y_0 \wedge \lnot x_0 \wedge \lnot x_1 \wedge \lnot y_1 \wedge \lnot y_2)) \wedge \\
            &\square((\var_\texttt{L2R} \wedge x_2 \wedge y_0 \wedge \lnot x_0 \wedge \lnot x_1 \wedge \lnot y_1 \wedge \lnot y_2) \rightarrow \\
            &\bigcirc (x_2 \wedge y_1 \wedge \lnot x_0 \wedge \lnot x_1 \wedge \lnot y_0 \wedge \lnot y_2)) \wedge \\
            &\square ((\var_\texttt{L2R} \wedge x_2 \wedge y_1 \wedge \lnot x_0 \wedge \lnot x_1 \wedge \lnot y_0 \wedge \lnot y_2) \rightarrow \\
            &\bigcirc (x_2 \wedge y_2 \wedge \lnot x_0 \wedge \lnot x_1 \wedge \lnot  y_0 \wedge \lnot y_1))
        \end{split}
    \end{equation}
Preconditions of \texttt{L2R} in $\syssafetynothard$ for Example~\ref{exm:9s} (Eq.~\ref{eq:sys_trans_choose},~\ref{eq:sys_trans_continue}): 
    \begin{equation}\label{eq:tsys_lefttoright}
    \begin{split}
        \syssafetychoose &\left\{
        \begin{array}{l}
            \square(\lnot(\bigcirc (x_0 \wedge y_0 \wedge \lnot x_1 \wedge \lnot x_2 \wedge \lnot y_1 \wedge \lnot y_2)\vee\\
            (x_0 \wedge y_0 \wedge \lnot x_1 \wedge \lnot x_2 \wedge \lnot y_1 \wedge \lnot y_2 \wedge \var_\texttt{L2R} \wedge \\
            \bigcirc (x_1 \wedge y_0 \wedge \lnot x_0 \wedge \lnot x_2 \wedge \lnot y_1 \wedge \lnot y_2)) \vee  \\
            (x_1 \wedge y_0 \wedge \lnot x_0 \wedge \lnot x_2 \wedge \lnot y_1 \wedge \lnot y_2 \wedge \var_\texttt{L2R} \wedge \\
            \bigcirc (x_2 \wedge y_0 \wedge \lnot x_0 \wedge \lnot x_1 \wedge \lnot y_1 \wedge \lnot y_2)) \vee  \\
            (x_2 \wedge y_0 \wedge \lnot x_0 \wedge \lnot x_1 \wedge \lnot y_1 \wedge \lnot y_2 \wedge \var_\texttt{L2R} \wedge \\
            \bigcirc (x_2 \wedge y_1 \wedge \lnot x_0 \wedge \lnot x_1 \wedge \lnot y_0 \wedge \lnot y_2)))\\
            \qquad\rightarrow \lnot \bigcirc \var_\texttt{L2R}) \wedge 
        \end{array}\right.\\
        \syssafetycontinue &\left\{
        \begin{array}{l}
        \square(x_0 \wedge y_0 \wedge \lnot x_1 \wedge \lnot x_2 \wedge \lnot y_1 \wedge \lnot y_2 \wedge \var_\texttt{L2R} \wedge\\
        \bigcirc (x_1 \wedge y_0 \wedge \lnot x_0 \wedge \lnot x_2 \wedge \lnot y_1 \wedge \lnot y_2) \rightarrow \\
        \bigcirc \var_\texttt{L2R}) \wedge \\
        \square (x_1 \wedge y_0 \wedge \lnot x_0 \wedge \lnot x_2 \wedge \lnot y_1 \wedge \lnot y_2 \wedge \var_\texttt{L2R} \wedge \\
        \bigcirc (x_2 \wedge y_0 \wedge \lnot x_0 \wedge \lnot x_1 \wedge \lnot y_1 \wedge \lnot y_2) \rightarrow \\
        \bigcirc \var_\texttt{L2R}) \wedge \\
        \square (x_2 \wedge y_0 \wedge \lnot x_0 \wedge \lnot x_1 \wedge \lnot y_1 \wedge \lnot y_2 \wedge \var_\texttt{L2R} \wedge \\
        \bigcirc (x_2 \wedge y_1 \wedge \lnot x_0 \wedge \lnot x_1 \wedge \lnot y_0 \wedge \lnot y_2) \rightarrow \\
        \bigcirc \var_\texttt{L2R})
        \end{array}
        \right.
    \end{split}
    \end{equation}
Symbolic suggestion for modifications to \texttt{L2R} in Example~\ref{exm:9s}:
    \begin{equation}\label{eq:nine_squares_suggestion}
    \begin{split}
        \square &( (x_0 \wedge y_0 \wedge \lnot x_1 \wedge \lnot x_2 \wedge \lnot y_1 \wedge \lnot y_2) \rightarrow \\
        &\bigcirc ((x_0 \wedge y_0 \wedge \lnot x_1 \wedge \lnot x_2 \wedge \lnot y_1 \wedge \lnot y_2) \vee \\
        &(x_1 \wedge y_0 \wedge \lnot x_0 \wedge \lnot x_2 \wedge \lnot y_1 \wedge \lnot y_2))) \wedge \\
        \square &( (x_1 \wedge y_0\wedge \lnot x_0 \wedge \lnot x_2 \wedge \lnot y_1 \wedge \lnot y_2) \rightarrow \\
        &\bigcirc ((x_1 \wedge y_0\wedge \lnot x_0 \wedge \lnot x_2 \wedge \lnot y_1 \wedge \lnot y_2) \vee \\
        &(x_1 \wedge y_1\wedge \lnot x_0 \wedge \lnot x_2 \wedge \lnot y_0 \wedge \lnot y_2)))\wedge \\
        \square &( (x_1 \wedge y_1\wedge \lnot x_0 \wedge \lnot x_2 \wedge \lnot y_0 \wedge \lnot y_2) \rightarrow \\
        &\bigcirc ((x_1 \wedge y_1\wedge \lnot x_0 \wedge \lnot x_2 \wedge \lnot y_0 \wedge \lnot y_2) \vee \\
        &(x_2 \wedge y_1\wedge \lnot x_0 \wedge \lnot x_1 \wedge \lnot y_0 \wedge \lnot y_2))) \wedge \\
        \square &( (x_2 \wedge y_1\wedge \lnot x_0 \wedge \lnot x_1 \wedge \lnot y_0 \wedge \lnot y_2) \rightarrow \\
        &\bigcirc ((x_2 \wedge y_1\wedge \lnot x_0 \wedge \lnot x_1 \wedge \lnot y_0 \wedge \lnot y_2) \vee \\
        &(x_2 \wedge y_2\wedge \lnot x_0 \wedge \lnot x_1 \wedge \lnot y_0 \wedge \lnot y_1)) \wedge \\
        \square &( (x_0 \wedge y_0\wedge \lnot x_1 \wedge \lnot x_2 \wedge \lnot y_1 \wedge \lnot y_2) \vee \\
        &(x_1 \wedge y_0\wedge \lnot x_0 \wedge \lnot x_2 \wedge \lnot y_1 \wedge \lnot y_2) \vee \\
        &(x_1 \wedge y_1\wedge \lnot x_0 \wedge \lnot x_2 \wedge \lnot y_0 \wedge \lnot y_2) \vee \\
        &(x_2 \wedge y_1\wedge \lnot x_0 \wedge \lnot x_1 \wedge \lnot y_0 \wedge \lnot y_2) \vee \\
        &(x_2 \wedge y_2\wedge \lnot x_0 \wedge \lnot x_1 \wedge \lnot y_0 \wedge \lnot y_1))
    \end{split}
\end{equation}
% \end{figure*}
% \newpage
% \newpage

\begin{algorithm*}[h]
\KwIn{$\tenv$, $\tenvhard$, $\tsysnothard$, $\tsyshard$, $Z$, $\inpuser$}
\KwOut{$\tenvnew$, $\tsysnewmutable$}
\BlankLine

\tcp{Find states where the system is guaranteed to reach $Z$ for any environment choice}
$\tsyscanwin = \{(\inpstateone, \outstateone, \inpstatetwo')\in \allstates \times \inpstatesprime \setor \exists \inpstateone, \inpstatetwo \in \inpstates, \outstateone, \outstatetwo \in \outstates \textrm{ s.t. } (\inpstateone, \outstateone, \inpstatetwo', \outstatetwo') \in \tsysnothard \textrm{ and } (\inpstatetwo, \outstatetwo) \in Z\}$\;\label{line:pre_syscanwin}

$\tsysalwayswins = \{(\inpstateone, \outstate) \in \allstates \setor \forall \inpstatetwo \in 2^{\inpsyms} \textrm{ s.t. } (\inpstateone, \outstate, \inpstatetwo') \in \tenv \rightarrow (\inpstateone, \outstate, \inpstatetwo') \in \tsyscanwin, \outstate \neq \emptyset\}$\;\label{line:pre_sysalwayswins}

$\treachable = \{(\inpstateone, \outstateone, \inpstatetwo') \in \tenv \setor \exists \inpstatethree \in \inpstates, \outstatethree \in \outstates \textrm{ s.t. } (\inpstatethree, \outstatethree, \inpstateone', \outstateone') \in \tsysnothard \textrm{ and } (\inpstatethree, \outstatethree, \inpstateone') \in \tenv$\}\;\label{line:pre_reachable}

$\tcanwinandencoded = \{ (\inpstateone, \outstateone, \inpstatetwo', \outstatetwo') \in \allstates \times  \allstatesprime \setor (\inpstateone, \outstateone) \in \tsysalwayswins,\ (\inpstatetwo, \outstatetwo) \in Z,\ (\inpstateone, \outstateone, \inpstatetwo') \in \treachable, (\inpstateone, \outstateone, \inpstatetwo', \outstatetwo') \in \tsysnothard\}$\;\label{line:pre_canwinandencoded}

\tcp{Find allowed changes to preconditions}
$\tau_\textrm{skills-changing} = \{\outstateone \in \outstates \setor \exists \inpstateone, \inpstatetwo \in \inpstates, \outstatetwo \in \outstates \textrm{ s.t. } (\inpstateone, \outstateone, \inpstateone', \inpstatetwo') \in \tcanwinandencoded\} $\;\label{line:pre_skillschanging}
$\tpreskills=\{(\inpstateone, \outstate) \in \allstates \setor \exists \inpstatetwo \in \inpstates \textrm{ s.t. } (\inpstateone, \outstate, \inpstatetwo) \in \treachable \textrm{ and } \outstate \in \tau_\textrm{skills-changing}\}$\;\label{line:pre_preskills}
$\tfinalpost=\{(\outstate, \inpstateone') \in \outstates \times \inpstatesprime \setor \exists \inpstatetwo \in \inpstates \textrm{ and } \nexists \inpstatethree \in \inpstates \textrm{ s.t. } (\inpstatetwo, \outstate, \inpstateone) \in \treachable \textrm{ and } (\inpstateone, \outstate, \inpstatethree) \in \treachable\}$\;\label{line:pre_finalpost}
$\tallowablechanges = \{ (\inpstateone, \outstateone, \inpstatetwo', \outstatetwo', \inpstatethree^{\dagger}) \in \allstates \times \allstatesprime \times \inpstatesdoubleprime \setor (\inpstateone, \outstateone, \inpstatetwo', \outstatetwo') \in \tcanwinandencoded,\ (\inpstateone, \inpstatethree') \in \tchangeconstraints,\ (\inpstatethree, \outstateone, \inpstatetwo') \in \tnotallowedrepair,\ (\inpstatethree, \outstateone) \notin \tpreskills,\ (\inpstatethree, \outstateone, \inpstatetwo', \outstatetwo') \in \tsyshard,\ (\inpstatethree, \outstateone, \inpstatetwo', \outstatetwo') \notin \tcanwinandencoded,\ \inpstatethree \neq \inpstatetwo,\ (\outstate, \inpstatethree') \notin \tfinalpost\}$\;\label{line:pre_allowablechanges}

\tcp{Randomly select one possible change to add}
\eIf{$\tallowablechanges \neq \emptyset$}{
$\tselectedchange = \textrm{randomlySelectOne}(\tallowablechanges)$\label{line:pre_selectedchange}\;
$\tselectedchange = \{(\inpstateone, \outstateone, \inpstatetwo', \outstatetwo', \inpstatethree^{\dagger})\in \allstates \times \allstatesprime \times \inpstatesdoubleprime \setor (\inpstateone, \outstateone, \inpstatetwo', \outstatetwo') \in \tsysnothard \textrm{ and } \exists \outstatethree \in \outstates \textrm{ s.t. } (\inpstateone, \outstateone, \inpstatetwo', \outstatethree', \inpstatethree^{\dagger}) \in \tselectedchange\}$\;\label{line:pre_selectadditionalchanges}
$\tselectedchange = \{(\inpstateone, \outstateone, \inpstatetwo', \outstatetwo', \inpstatethree^{\dagger}) \in \allstates \times \allstatesprime \times \inpstatesdoubleprime \setor (\inpstateonea, \outstateone, \inpstatetwoa', \outstatetwo, \inpstatethreea^{\dagger}) \in \tselectedchange, \statevar_\textrm{u} \in 2^{\inpuser}, \inpstateone = (\inpstateonea \setminus \inpuser) \cup \statevar_\textrm{u}, \inpstatetwo = (\inpstatetwoa \setminus \inpuser) \cup \statevar_\textrm{u}, \inpstatethree = (\inpstatethreea \setminus \inpuser) \cup \statevar_\textrm{u}\}$\;\label{line:pre_alluservariables}
}
{
\Return $\tenv$, $\tsysnothard$\
}
$\tnewfullskill = \{(\inpstatethree, \outstateone, \inpstatetwo', \outstatetwo') \in \allstates \times \allstatesprime \setor (\inpstateone, \outstateone, \inpstatetwo', \outstatetwo', \inpstatethree^{\dagger}) \in \tselectedchange \}$\;\label{line:pre_newfullskill}

\tcp{Add transitions to $\tsysnothard$}

$\toldfullskill = \{(\inpstateone, \outstateone, \inpstatetwo', \outstatetwo') \in \allstates \times \allstatesprime \setor (\inpstateone, \outstateone, \inpstatetwo', \outstatetwo', \inpstatethree^{\dagger}) \in \tselectedchange \}$\;\label{line:pre_oldfullskill}

$\toldprepreconditions = \{\inpstateone, \outstateone) \in \allstates \setor (\inpstatetwo, \outstatetwo, \inpstatethree, \outstatethree) \in \toldfullskill \textrm{ and } (\inpstateone, \outstateone, \inpstatetwo', \outstatetwo') \in \tsysnothard\}$\;\label{line:pre_oldprepreconditions}
$\tnewpreprecondition = \{(\inpstateone, \outstateone, \inpstatetwo', \outstatetwo') \in \allstates \times \allstatesprime \setor (\inpstateone, \outstateone) \in \toldprepreconditions \textrm{ and } (\inpstatetwo, \outstatetwo, \inpstatethree, \outstatethree) \in \tnewfullskill\}$\;\label{line:pre_newprepreconditions}

\tcp{Determine if this is the initial precondition}
\If{$\{(\inpstateone, \outstateone, \inpstatetwo', \outstatetwo') \in \allstates \times \allstatesprime \setor \forall \outstateone \in \outstates, (\inpstateone, \outstateone, \inpstatetwo', \outstatetwo') \in \tnewpreprecondition\} == \tnewpreprecondition$}{\label{line:pre_modinitprecheck} 
$\tau_\textrm{only-input-props} = \{\inpstatetwo \in \inpstate \setor (\inpstateone, \outstateone, \inpstatetwo', \outstatetwo') \in \tnewpreprecondition\}$\;
$\tau_\textrm{diff-skills-with-same-pre} = \{(\inpstateone, \inpstatetwo, \outstatetwo) \in \inpstates \times \allstates \setor \forall \outstateone \in \outstate, (\inpstateone, \outstateone, \inpstatetwo, \outstatetwo) \in \tsysnothard \textrm{ and } \inpstatetwo \in \tau_\textrm{only-input-props}\}$\;\label{line:pre_allskillscanbeapplied}
$\tau_\textrm{remove} = \{(\inpstateone, \outstateone, \inpstatetwo') \in \allstates \times \inpstatesprime \setor \exists \outstatetwo \in \outstates \textrm{ s.t. } (\inpstateone, \outstateone, \inpstatetwo, \outstatetwo) \in \tsysnothard \textrm{ and } (\inpstateone, \inpstatetwo, \outstatetwo) \notin \tau_\textrm{diff-skills-with-same-pre} \textrm{ and } \inpstatetwo \in \tau_\textrm{only-input-props}\}$\;\label{line:pre_remove}
$\tnewpreprecondition = \{(\inpstateone, \outstateone, \inpstatetwo, \outstatetwo) \in \tnewpreprecondition \setor (\inpstateone, \outstateone, \inpstatetwo') \notin \tau_\textrm{remove}\}$\;\label{line:pre_newprepreconditions2}
}
$\tsysnewmutable = \tsysnothard \cup \tnewpreprecondition \cup \tnewfullskill$\;\label{line:pre_sysnothardnew}

\tcp{Add transitions to $\tenv$}

$\tenvpreold = \{(\inpstateone, \outstateone) \in \allstates\setor (\inpstateone, \outstateone, \inpstatetwo') \in\treachable \textrm{ and } (\inpstatetwo, \outstateone, \inpstatethree', \outstatethree') \in \toldfullskill\}$\;\label{line:pre_envpreold}
$\tenvprenew = \{(\inpstateone, \outstateone, \inpstatetwo') \in \allstates \times \inpstatesprime \setor (\inpstateone, \outstateone) \in \tenvpreold,\ (\inpstatetwo, \outstateone, \inpstatethree', \outstatethree') \in \tnewfullskill\}$\;\label{line:pre_envprenew}
$\tenvnewfullskill = \{(\inpstateone, \outstateone, \inpstatetwo') \in \allstates \times \inpstatesprime \setor (\inpstateone, \outstateone, \inpstatetwo', \outstatetwo') \in \tnewfullskill\}$\;\label{line:pre_envnewfullskill}
$\tenvnew = (\{(\inpstateone, \outstateone, \inpstatetwo') \in \tenv \setor \nexists \inpstatethree, \outstatethree \textrm{ s.t. } (\inpstateone, \outstateone, \inpstatethree', \outstatethree') \in \tnewfullskill\} \cup \tenvnewfullskill \cup \tenvprenew) \cap \tenvhard$\;\label{line:pre_envnew2}
\Return $\tenvnew$, $\tsysnewmutable$
\caption{\textbf{modify\_preconditions}}
\label{alg:modifyPre}
\end{algorithm*}

\begin{figure*}
\removelatexerror
\input{symbolicalgorithms/findSimilarPostconditions}
\vspace*{4.8in}
\end{figure*}

\end{document}